%% file: main.tex
\documentclass[runningheads]{llncs}

\usepackage{eccv}

\usepackage{eccvabbrv}

\usepackage{graphicx}
\usepackage{booktabs}

\usepackage[accsupp]{axessibility}  % Improves PDF readability for those with disabilities.

\def\etal{{\textit{et~al.}}}
\def\fig{{Fig.}}

\def\tab{{Tab.}}
\def\eq{{Eq.}}
\def\sec{{Sec.}}
\def\alg{{Algorithm}}
\def\oursdataset{{RefracGS}}

\definecolor{vscodegreen}{HTML}{6A9955}
\newcommand{\codecomment}[1]{\textcolor{vscodegreen}{// #1}}

\usepackage{hyperref}

\usepackage{orcidlink}
\input{sec/notation}

\usepackage{verbatim}
\usepackage{cuted}
\usepackage{lipsum}
\usepackage{bm}
\usepackage{comment}
\newlength{\imagewidth}
\usepackage{float}
\usepackage{multirow}
\usepackage{tabularx}
\usepackage{algorithm}
\usepackage{algpseudocode}
\usepackage{mathrsfs}
\usepackage{tikz}
\usetikzlibrary{spy}

\usepackage{subcaption}
\usepackage{caption}
\usepackage{wrapfig}

\usepackage{url}

\usepackage{xcolor}
\usepackage[table]{xcolor}
\definecolor{link_color}{RGB}{0, 0, 128}
\definecolor{table_highlight}{RGB}{166,202,164}
\definecolor{firstscore}{RGB}{241,158,156}
\definecolor{secondscore}{RGB}{247,206,160}
\definecolor{thirdscore}{RGB}{253,248,182}

\begin{document}

\title{RefracGS: Novel View Synthesis Through Refractive Water Surfaces with 3D Gaussian \\ Ray Tracing}

\titlerunning{RefracGS}

% Second Author\inst{2,3}\orcidlink{1111-2222-3333-4444} \and
% Third Author\inst{3}\orcidlink{2222--3333-4444-5555}}
\author{
Yiming Shao\inst{1}\thanks{Equal contribution. \protect\newline \makebox[-0.35em][r]{$^{\dagger}$}~Corresponding authors.} \and
Qiyu Dai\inst{2}$^{\star}$ \and
Chong Gao\inst{3} \and
Guanbin Li\inst{3, 5} \and
Yequan Wang\inst{2, 4} \and \\
He Sun\inst{2} \and
Qiong Zeng\inst{1}$^{\dagger}$ \and
Baoquan Chen\inst{2} \and
Wenzheng Chen\inst{2, 4}$^{\dagger}$
}

\authorrunning{Y.~Shao et al.}

% Springer Heidelberg, Tiergartenstr.~17, 69121 Heidelberg, Germany
\institute{Shandong University, China \and
Peking University, Beijing, China \and
Sun Yat-sen University, China \and
Beijing Academy of Artificial Intelligence, Beijing, China \and
Shenzhen Loop Area Institute, China}

\input{figures/tex/teaser}

\input{sec/0_abstract}  
% \begin{abstract}
%   The abstract should concisely summarize the contents of the paper. 
%   While there is no fixed length restriction for the abstract, it is recommended to limit your abstract to approximately 150 words.
%   Please include keywords as in the example below. 
%   This is required for papers in LNCS proceedings.
%   \keywords{First keyword \and Second keyword \and Third keyword}
% \end{abstract}

\input{sec/1_intro}
\input{sec/2_related_work}

\input{sec/3_methodology_v2}

\input{sec/4_experiments}
\input{sec/5_conclusion}

\bibliographystyle{splncs04}
\bibliography{main}

\input{sec/X_suppl}
\end{document}

%% file: sec/notation.tex
\newcommand{\name}{RefracGS}

\newcommand{\watersurface}{Water Height Map}

%% file: figures/tex/teaser.tex
\maketitle
\begin{center}
    \centering
    \includegraphics[width=\textwidth]{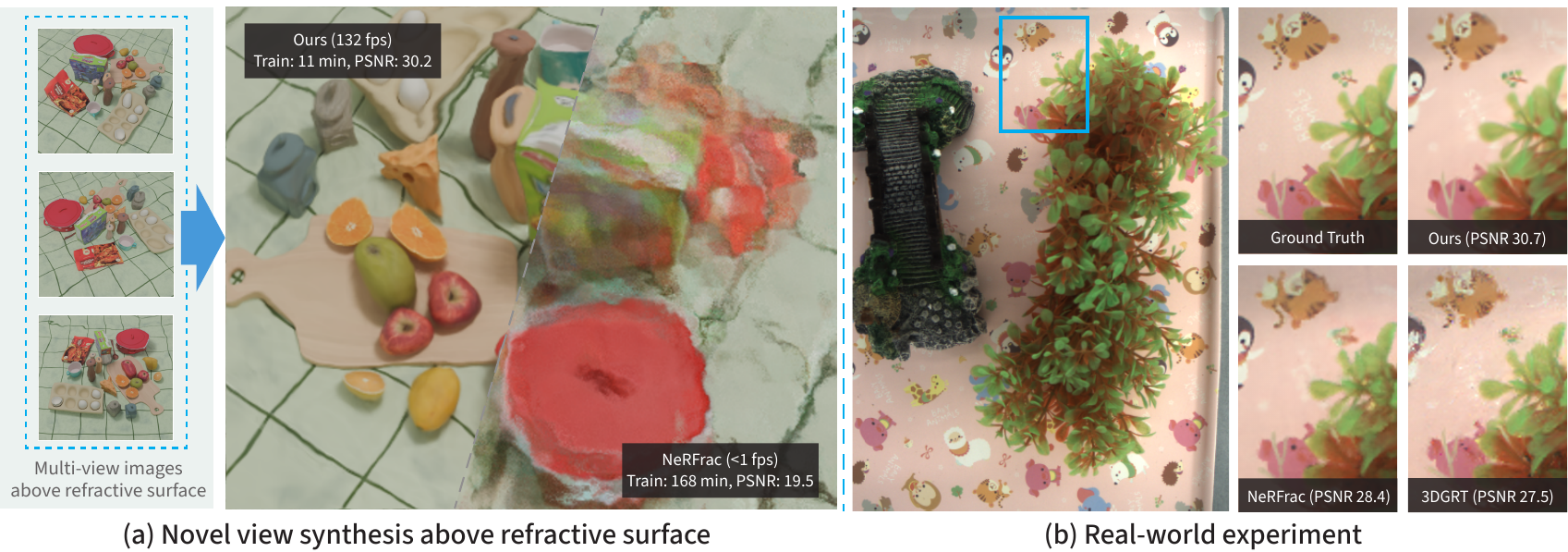}
    \vspace{-6mm}
    \captionof{figure}{\textbf{(a):} Our method reconstructs both the refractive water surface and the underlying scene from multi-view images captured above, achieving high-fidelity novel view synthesis with fast training and real-time rendering. \textbf{(b):} Experiments in real-world scenarios further demonstrate the effectiveness of our approach.}
    \label{fig:teaser}
\end{center}

%% file: sec/0_abstract.tex
\begin{abstract}
Novel view synthesis (NVS) through non-planar refractive surfaces presents fundamental challenges due to severe, spatially varying optical distortions. While recent representations like NeRF and 3D Gaussian Splatting (3DGS) excel at NVS, their assumption of straight-line ray propagation fails under these conditions, leading to significant artifacts.
To overcome this limitation, we introduce RefracGS, a framework that jointly reconstructs the refractive water surface and the scene beneath the interface. Our key insight is to explicitly decouple the refractive boundary from the target objects: the refractive surface is modeled via a neural height field, capturing wave geometry, while the underlying scene is represented as a 3D Gaussian field. We formulate a refraction-aware Gaussian ray tracing approach that accurately computes non-linear ray trajectories using Snell's law and efficiently renders the underlying Gaussian field while backpropagating the loss gradients to the parameterized refractive surface. Through end-to-end joint optimization of both representations, our method ensures high-fidelity NVS and view-consistent surface recovery. Experiments on both synthetic and real-world scenes with complex waves demonstrate that RefracGS outperforms prior refractive methods in visual quality, while achieving \textbf{$\sim$15$\times$ faster} training and real-time rendering at \textbf{200 FPS}.
\keywords{Novel view synthesis \and Refracion \and Ray tracing}
\end{abstract}

%% file: sec/1_intro.tex
\vspace*{-4mm}
\section{Introduction}
\label{sec:intro}

Novel view synthesis (NVS) of scenes observed through non-planar refractive surfaces (e.g., water surfaces) presents a long-standing and fundamentally challenging problem in computer vision.
Such capabilities hold significant promise for a wide range of real-world applications, including shallow water bathymetry, environmental monitoring, and underwater cultural heritage preservation.
In contrast to NVS in air, NVS through refractive surfaces tackles complex distortions and disrupted correspondences caused by the refractive effects of the interface. 
As light rays traverse interfaces between media with mismatched refractive indices, they undergo abrupt changes in direction, leading to severe optical distortions, particularly in the presence of surface undulations.
This refraction-induced bending fundamentally invalidates the multi-view consistency and linear projection assumptions central to standard reconstruction, and effective solutions to this ill-posed problem remain largely absent in existing work.

Recent advances in Neural Radiance Fields (NeRF)~\cite{mildenhall2020nerf} and 3D Gaussian Splatting (3DGS)~\cite{kerbl3Dgaussians} have enabled high-fidelity 
NVS for in-air scenes.
However, these methods predominantly assume linear radiance transport within a homogeneous medium and thus fail under refractive conditions.
To address non-linear light paths introduced by refraction, prior works have attempted explicit modeling. NeRF-based methods, such as NeRFrac~\cite{zhan2023nerfrac}, incorporate ray-bending to simulate refractive effects, but suffer from the high computational cost of volumetric rendering, limiting both training and inference speed.
More crucially, many such approaches define the refractive interface in view-dependent spaces—e.g., parameterizing surface normals as functions of camera rays
—which introduces geometric inconsistencies across views and hampers robust surface reconstruction~\cite{zhan2023nerfrac,wang2023neref}.
Parallel efforts have started extending 3DGS to support non-linear light transport, as seen in TransparentGS~\cite{huang2025transparentgs}, which focuses on the reconstruction of transparent solid objects. Yet, these methods are not designed for through-surface imaging, where refractive distortion arises from a dynamic fluid interface.

In this work, we propose {\name}, a novel framework that simultaneously reconstructs both the refractive water surface geometry and the underlying scene by coupling a novel refractive surface representation with 3D Gaussian ray tracing.
Specifically, we introduce {\watersurface}, a hybrid refractive surface representation that combines a contiguous neural height field with a recursively subdivided triangular mesh. 
This design provides mesh-like efficiency for ray–surface intersection queries and field-like continuity for accurate, view-consistent surface geometry.

Built upon {\watersurface}, we develop a refraction-aware Gaussian ray tracing algorithm that explicitly enforces Snell’s law at the air–water interface. 
For each camera pixel, we first trace a ray to intersect the reconstructed refractive surface using a hierarchical subdivision scheme, then bend the ray according to the estimated local surface normal, and finally accumulate color and opacity along the refracted path through a set of 3D Gaussian primitives modeling the underlying scene. 
Different from 3DGRT~\cite{loccoz20243dgrt}, our entire pipeline is fully differentiable: gradients flow not only to the Gaussian parameters (positions, opacities, and appearance coefficients) but also to the parameterized refractive surface, enabling joint optimization from image supervision alone. To the best of our knowledge, this is the first method to directly optimize refractive surfaces using Gaussian ray tracing.
Implemented with efficient CUDA kernels and hardware-accelerated ray traversal, our method preserves much of the speed advantage of Gaussian Splatting while faithfully modeling physical refraction.

We conduct extensive experiments on both synthetic and real-world scenes with complex wave dynamics. The results show that RefracGS consistently outperforms all baselines, achieving state-of-the-art performance in both visual quality and surface geometry accuracy. In addition, our method is approximately \textbf{$\sim$15$\times$} faster to train while supporting real-time rendering at over \textbf{200 FPS}. Finally, the explicit disentanglement of the refractive surface extends beyond novel view synthesis, enabling flexible and controllable scene manipulation, such as water removal and surface editing, without retraining.

%% file: sec/2_related_work.tex
\vspace{-4mm}
\section{Related Work}
\label{sec:related-work}

\vspace{-2mm}
In this section, we briefly review two related topics: recent advances in novel view synthesis, and methods for reconstructing refractive surfaces. 

\vspace{-2mm}
\paragraph{Novel View Synthesis}
Novel view synthesis (NVS) has witnessed remarkable advancements in recent years. NeRF~\cite{mildenhall2020nerf} models scenes with a multilayer perceptron (MLP) and synthesizes new views via volumetric rendering. Numerous subsequent works have aimed to enhance the computational efficiency of NeRF~\cite{yu2022plenoxels,Chen2022tensorf,mueller2022instant}, as well as the capability of handling aliasing~\cite{barron2021mipnerf}, unbounded scene~\cite{barron2022mipnerf360}, and other challenges.

Instead of using neural network to represent scenes, 3D Gaussian Splatting~\cite{kerbl3Dgaussians} uses 3D Gaussian particles, which can be rendered through highly efficient rasterization, achieving competitive results in terms of both quality and efficiency. Several follow-up studies have proposed improvements to 3D Gaussian Splatting~\cite{ye2024absgs,kheradmand20243dgsmcmc,lee2024c3dgs,fan2023lightgaussian}. In addition, some works have focused on extracting high-quality geometry from Gaussian representations~\cite{Huang2DGS2024,Yu2024GOF,chen2024pgsr}.
Although the rasterization-based rendering in 3D Gaussian Splatting enables extremely fast speed, it inherently limits the method’s ability to handle complex light transport phenomena such as reflection and refraction. Some methods~\cite{jiang2023gaussianshader,yao2025refGS} represent reflective objects through material decomposition and related techniques, but they still lack explicit tracing of light paths and are not applicable to the reconstruction of refracted objects. 3DGRT~\cite{loccoz20243dgrt} replaces rasterization with ray tracing for rendering, allowing the simulation of complex light transport phenomena. While this approach inspires our work, the direct reconstruction of refractive objects using Gaussian ray tracing remains largely unexplored.

\vspace{-3mm}
\paragraph{Refractive Surface Reconstruction}
Reconstructing refractive surfaces and the objects behind them has long been a challenging task. Several approaches~\cite{bemana2022eikonal,pan2022sample,kim2023ref2nerf,Deng2024RayDeformation} have augmented NeRF to support the modeling of refractive objects. Other methods~\cite{wang2023nemto,Tong2023seeing,gao2023transparent,Jia2024NUNeRF} adopt Signed Distance Function (SDF) to model refractive surfaces. Since NeRF and SDF based methods adopt ray-based rendering, they are naturally suited to handling the light path bending caused by refraction. However, the rasterization-based rendering in 3D Gaussian Splatting makes it difficult to handle refraction phenomena. TransparentGS~\cite{huang2025transparentgs} bakes refracted background content using light field probes for transparent object reconstruction. But this method approximates the Gaussian radiance field via light field probes, inherently introducing deviations relative to accurate ray tracing. Crucially, these approaches are primarily designed for reconstructing isolated transparent objects, rendering them ill-equipped for through-surface imaging where refractions arise from a continuous, non-planar interface.

In addition, several studies aim to reconstruct both the refractive water surface and the underwater scene from images captured above water. Qian \etal~\cite{qian2018simultaneous} reconstruct both the water surface and the underwater scene by leveraging multi-view consistency and variational optical flow estimation. Xiong \etal~\cite{xiong2021inthewild} reconstruct both the water surface and the underwater scene from a single-camera video sequence. NeReF~\cite{wang2023neref} reconstructs fluid surfaces from multi-view images, but it assumes access to a known underwater pattern, limiting its applicability in real-world settings. More recently, WSGS~\cite{yoon2026wsgs} addresses refractive underwater reconstruction; however, it relies on an idealized assumption of a perfectly planar water surface. NeRFrac~\cite{zhan2023nerfrac} simultaneously reconstructs non-planar refractive surfaces and underlying scenes from multi-view images, providing inspiration for our work.

%% file: sec/3_methodology_v2.tex
\begin{figure*}[t]
    \vspace{-2mm}
	\centering
	\includegraphics[width=1\linewidth]{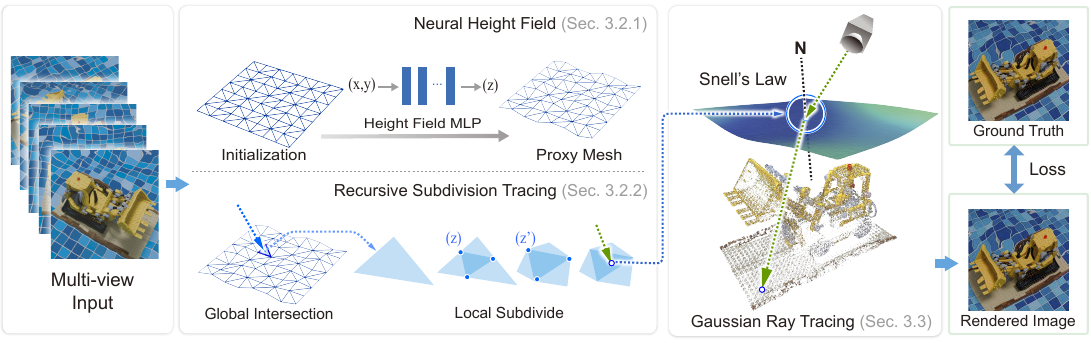}
	\vspace{-6mm}
    \caption{
    \label{fig:pipeline} 
    \textbf{Overview of the {\name} pipeline}. A proxy refractive‑surface mesh is first extracted from the neural height field, after which camera rays intersect the mesh using a hierarchical global‑to‑local strategy to obtain intersection points and normals. The refracted directions are then computed with Snell’s law, and the resulting rays traverse the underlying scene before being rendered with Gaussian ray tracing.
    }
    \vspace{-10mm}
\end{figure*}

\section{Methodology}
\label{sec:methodology}

Given a series of images captured from multiple viewpoints above a refractive surface, along with their corresponding camera parameters, our method aims to reconstruct both the Refractive surface geometry and the underlying scene. The overall pipeline of our proposed {\name} framework is illustrated in \fig~\ref{fig:pipeline}. The core of our approach lies in two key components: (1) a {\watersurface} for accurate and differentiable modeling of the refractive surface, and (2) a tailored refraction-aware ray tracing mechanism compatible with 3D Gaussian Splatting~\cite{kerbl3Dgaussians} (3DGS). In \sec~\ref{sec:method:preliminaries}, we first review 3D Gaussian Splatting and the physical law of refraction, followed by a comprehensive explanation on our {\watersurface} (\sec~\ref{sec:method:surface}) and the refraction-aware Gaussian ray tracing algorithm (\sec~\ref{sec:method:grt}). Implementation details including loss function and optimization hyper-parameters are detailed in \sec~\ref{sec:method:impl} to further enhance the reproducibility.

\subsection{Preliminaries}
\label{sec:method:preliminaries}
\vspace{-1mm}
\paragraph{3D Gaussian Splatting}
3D Gaussian Splatting~\cite{kerbl3Dgaussians} represents a scene as a collection of anisotropic 3D Gaussian primitives 
$\mathcal{G} = \{ \mathcal{G}_i \}_{i=1}^N$, where each Gaussian $\mathcal{G}_i$ is defined by its central position 
$\bm{\mu}_i \in \mathbb{R}^3$, covariance matrix $\bm{\Sigma}_i \in \mathbb{R}^{3\times3}$, 
opacity $o_i \in [0,1]$, color $\bm{c}_i \in [0,1]^3$, and optionally spherical harmonic coefficients. 
Each Gaussian primitive models a volumetric density distribution:
\vspace{-1mm}
\begin{equation}
\alpha_i(\bm{x}) = o_i \, \exp\left(-\frac{1}{2} (\bm{x} - \bm{\mu}_i)^\top 
\bm{\Sigma}_i^{-1} (\bm{x} - \bm{\mu}_i) \right),
\end{equation}
where $\bm{\Sigma}_i=\textbf{R}_i\textbf{S}_i\textbf{S}_i^\top\textbf{R}_i^\top$ is decomposed into a rotation matrix $\textbf{R}_i\in \mathbb{R}^{3\times3}$ and a scale matrix $\textbf{S}_i\in \mathbb{R}^{3\times3}$ to ensure the positive semi-definite nature of the covariance matrix.
During rendering, an efficient rasterization is conducted to project each Gaussian onto the image plane,
resulting in a 2D elliptical footprint. The final color $\textbf{C}$ of pixel $\bm{p}$ is computed by alpha compositing these projected 
Gaussians in front-to-back order:
\vspace{-2mm}
\begin{equation}
\textbf{C} = \sum_{i=1}^{N} T_i(\bm{p})\, \alpha_i(\bm{p}) \, \mathbf{c}_i \,,
\end{equation}
where $T_i(\mathbf{p}) = \prod_{j < i}\left( 1 - \alpha_j(\mathbf{p})\right)\,$ denotes transmittance.

\paragraph{3D Gaussian Ray Tracing}
Despite the remarkable efficiency of the rasterization process, it inherently assumes linear radiance propagation, an assumption that often fails in complex real-world scenarios involving reflective surfaces or refractive interfaces. To address this limitation, 3DGRT~\cite{loccoz20243dgrt} introduces a ray tracing–based rendering technique specifically designed for 3DGS, enabling the modeling of directional deviations caused by reflection and refraction while maintaining real-time performance. However, the proposed ray tracing formulation is not differentiable with respect to the reflective or refractive properties of surfaces, and thus remains incapable of reconstructing objects with unknown reflection or refraction characteristics.

\vspace{-3mm}
\paragraph{The Law of Refraction}
In nature, the refraction of light is governed by Snell’s law, which describes how a ray of light changes direction when crossing media boundaries:
\vspace{-1mm}
\begin{equation}
n_1 \sin \theta_1 = n_2 \sin \theta_2,
\end{equation}
where $n_1$ and $n_2$ are the refractive indices of the two media, and $\theta_1$ and $\theta_2$ denote the angles of incidence and refraction, respectively.
For computational convenience, we adopt the vector formulation of Snell’s law:
\vspace{-1mm}
\begin{equation}
\mathbf{\omega_o} = \eta (\mathbf{\omega_i} + c_1 \mathbf{n}) - c_2 \mathbf{n},
\label{eq:snell:vector}
\end{equation}
where $\mathbf{\omega_i} \in \texttt{SO(3)}$ is the incident ray direction, $\mathbf{n} \in \texttt{SO(3)}$ is the surface normal, and $\mathbf{\omega_o} \in \texttt{SO(3)}$ is the refracted ray direction. Here, $\eta = {n_1}/{n_2}$ is the ratio of refractive indices, $c_1 = \mathbf{n} \cdot \mathbf{\omega_i}$, and $c_2 = \sqrt{1 - \eta^2 (1 - c_1^2)}$.

\subsection{\watersurface}
\label{sec:method:surface}
\vspace{-1mm}
\paragraph{Overview}
A precise and differentiable representation of the Refractive surface is crucial for accurately simulating refraction.
For any given ray defined by an origin $\bm{o} \in \mathbb{R}^3$ and direction $\bm{d} \in \texttt{SO(3)}$, our goal is to determine two key quantities: the intersection point $\bm{p} \in \mathbb{R}^3$ on the surface where refraction occurs, and the resulting refracted ray direction $\mathbf{\omega_o} \in \texttt{SO(3)}$ as it propagates into the underlying scene.
Therefore, an effective refractive surface model $\mathcal{W}$ must provide the intersection depth $t\in\mathbb{R}^+$ for intersection position computation $\bm{p}=\bm o+t\bm d$ and the surface normal $\mathbf{n}\in\texttt{SO(3)}$ at $\bm{p}$ for refracted direction computation, as formulated in Eq.~\ref{eq:water:surface}:
\begin{equation}
    (t,\mathbf{n})=\mathcal{W}(\bm{o},\bm{d}) \,.
    \label{eq:water:surface}
\end{equation}
Counterintuitively, modeling a refractive surface both accurately and efficiently remains a challenge.

\begin{figure}[t]
  \centering
  \begin{minipage}[b]{0.48\textwidth}
    \centering
    \setlength{\tabcolsep}{3pt}
    \renewcommand{\arraystretch}{1.0}
    \begin{tabular}{cc}
      \includegraphics[width=0.38\linewidth]{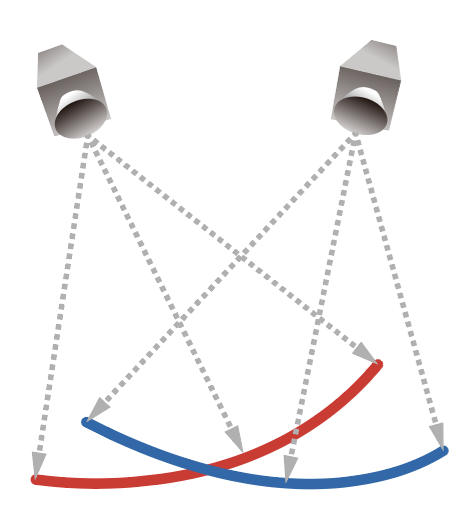} &
      \includegraphics[width=0.5\linewidth]{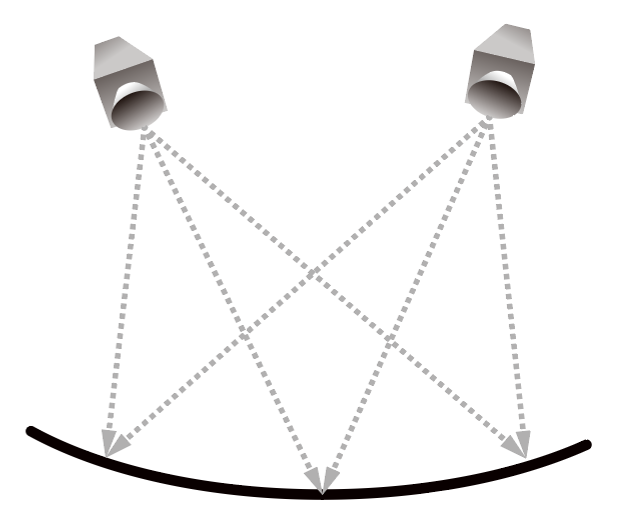} \\
      \small (a) view inconsistent & \small (b) view consistent
    \end{tabular}
    \vspace{-1mm} 
    \caption{\textbf{Surface modeling strategy.} NeRFrac (a) predicts ray–surface intersections using an MLP conditioned on ray origin and direction, which fails to enforce multi-view geometric consistency. In contrast, our method (b) explicitly models the refractive surface in world space, ensuring consistent intersection positions across viewpoints.}
    \label{fig:surface_modeling_consis}
  \end{minipage}
  \hfill
  \begin{minipage}[b]{0.48\textwidth}
    \centering
    \setlength{\tabcolsep}{3pt}
    \renewcommand{\arraystretch}{1.0}
    \begin{tabular}{cc}
      \includegraphics[width=0.47\linewidth]{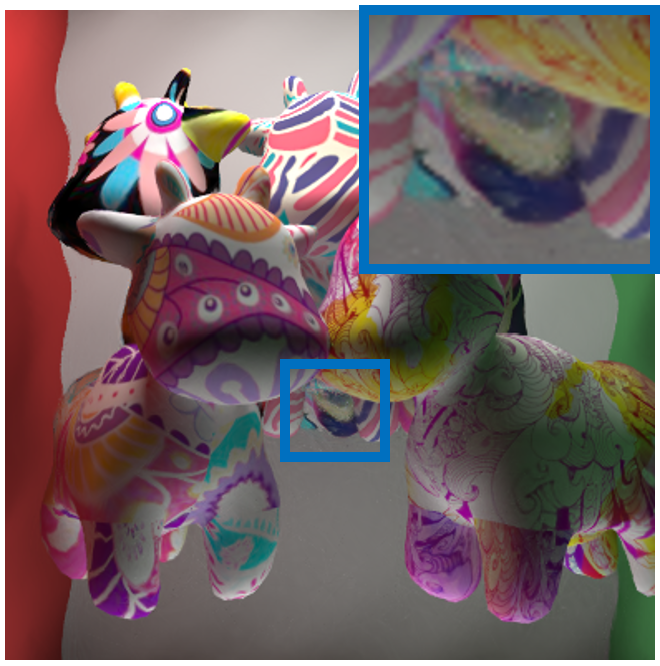} &
      \includegraphics[width=0.47\linewidth]{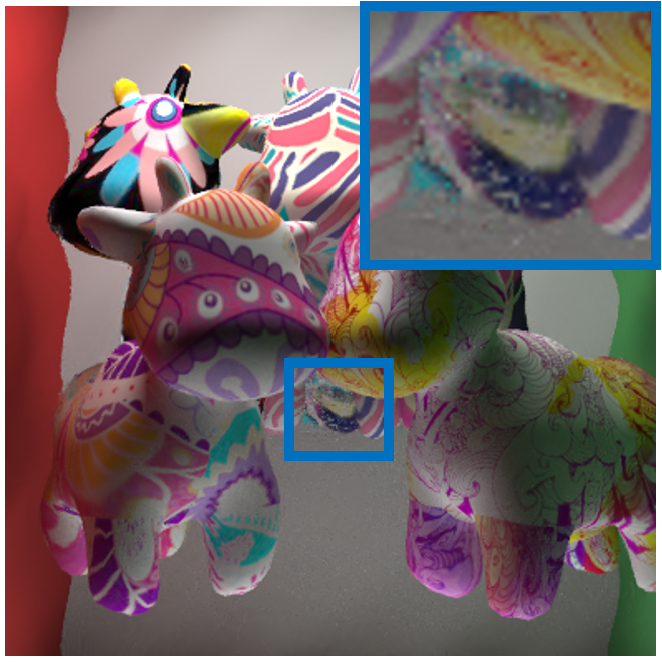} \\
      \small (c) w opacity loss & \small (d) w/o opacity loss
    \end{tabular}
    \vspace{-1mm}
    \caption{\textbf{Effectiveness of opacity loss.} We compare rendering results (c) with and (d) without our proposed opacity loss. The visualization demonstrates that applying this regularization effectively eliminates floating artifacts in complexly occluded regions, yielding significantly cleaner scene reconstructions.}
    \label{fig:opacity_loss_effect}
  \end{minipage}
\end{figure}

Existing representations for refractive surfaces, such as explicit meshes, struggle to model high-frequency geometry like water waves due to limited resolution. Alternatively, implicit field methods like NeRFrac~\cite{zhan2023nerfrac} leverage MLPs to model contiguous surfaces but suffer from view-dependent queries that lack multi-view consistency (\fig~\ref{fig:surface_modeling_consis}) and require the computationally expensive plane fitting to estimate normal directions from the implicitly represented surfaces. To address these limitations, we propose {\watersurface}, a novel representation that models the interface as a neural height field. By querying intersections via recursive subdivision tracing, our method combines the efficient intersection queries of meshes with the contiguous geometry modeling of fields, inherently achieving both spatial consistency and computational efficiency.

\begin{comment}
    
\end{comment}

\vspace{-1mm}
\subsubsection{Neural Height Field}
\label{sec:method:height}

Following NeRFrac~\cite{zhan2023nerfrac}, we define the $Z$-axis as the height direction, pointing upward from the underlying medium. This allows the refractive surface—approximately perpendicular to the $Z$-axis and aligned with the $XY$-plane—can be represented as a height field $\mathcal{H}(x, y)$:
\vspace{-1mm}
\begin{equation}
\mathcal{H}(x, y) = z\,,
\end{equation}
where $\mathcal{H}(x, y) - z =0$ defines an implicit function that formulates the refractive surface as its zero isosurface. The height function $\mathcal{H}(x, y)$ is parameterized by a multi-layer perceptron (MLP), providing a continuous, smooth, and view-independent representation of the surface geometry. Compared to NeRFrac's view-dependent refractive field, this formulation improves spatial consistency, facilitating stable multi-view optimization. 

To model refraction, the height field should derive the intersection depth $t$ and surface normal $\mathbf{n}$ for each ray query, as formulated in Eq.~\ref{eq:water:surface}. However, directly computing intersections with the implicit MLP-defined isosurface is computationally expensive. To enable efficient geometric queries, we propose the Recursive Subdivision Tracing tailored for our Neural Height Field, as detailed in the following section.

\subsubsection{Recursive Subdivision Tracing}
\label{sec:method:subdivision}

Given the ray query $\bm{r}=\bm{o}+t\bm{d}$ on a neural height field $\mathcal{H}$, our goal is to efficiently solve $t$, so that
\begin{equation}
    \mathcal{H}(o_x+td_x,o_y+td_y)=o_z+td_z \,.
    \label{eq:method:intersection:solution}
\end{equation}
However, as discussed in \sec~\ref{sec:method:height}, the implicit function nature of $\mathcal{H}$ hinders a direct solution of Eq.~\ref{eq:method:intersection:solution}, and the conventional ray marching-based binary search is computationally impractical, especially when conducted during each rendering process.

\paragraph{Proxy-based Tracing} To this end, we employ a proxy mesh surface as a piecewise linear approximation of $\mathcal{H}$ for accelerated ray-surface intersection. As illustrated in \fig~\ref{fig:pipeline} (1), we predefine a 2D triangular mesh $\mathcal{M_\textrm{2D}}(\mathcal{V}_\textrm{2D},\mathcal{F})$ with $\mathcal{V}_\textrm{2D}=\{(x_i,y_i)\in\mathbb{R}^2\}_{i=1}^M$ denoting its vertex set and $\mathcal{F}$ representing the triangular connectivity. For each 2D vertex $(x_i,y_i)$, we lift it to 3D by querying its height $z_i=\mathcal{H}(x_i,y_i)$. This results in a 3D proxy mesh serving as the piecewise linear approximation of $\mathcal{H}$, thereby transforming the ray–height field intersection problem into a more tractable ray–mesh intersection problem, where the computation naturally benefits from modern hardware (\textit{e.g.}, RT Cores) and can be accelerated by Bounding Volume Hierarchies (BVHs).

\paragraph{Recursive Subdivision} Despite the efficiency, the limited resolution of the mesh proxy hinders accurate refraction modeling when confronted with complex interface involving detailed surface geometry. To this end, we propose a coarse-to-fine strategy for progressive approximation of $\mathcal{H}$. As outlined in \alg~\ref{alg:tracing}, we introduce the Recursive Subdivision Tracing, where $\mathcal{M_\textrm{2D}}$ will be recursively subdivided and the newly added vertices will updated their heights via $\mathcal{H}$ queries. Furthermore, this tracing approach can fully leverage locality by gradually pruning non-intersected triangles, eliminating the need for repeated BVH construction and bypassing redundant MLP evaluations. As a result, our proposed Recursive Subdivision Tracing achieves excellent efficiency while benefiting from great surface modeling ability of the neural height field.

\subsubsection{Normal Estimation}
\label{sec:method:normal}
In this section, we detail the calculation of the normal direction in Algorithm~\ref{alg:tracing}. Given the ray–mesh intersection point $\bm{p}$ and the hit triangle facet $\mathcal{F}=(i,j,k)$, a straightforward strategy is to use the facet normal $\bm{n}_{\mathcal{F}}=\frac{(\bm{v}_j-\bm{v}_i)\times(\bm{v}_k-\bm{v}_i)}{\|(\bm{v}_j-\bm{v}_i)\times(\bm{v}_k-\bm{v}_i)\|_2}$ for refraction computation in Eq.~\ref{eq:snell:vector}. However, these face normals create discontinuity at triangle boundaries, adversely degrading rendering quality.

Inspired by Phong shading~\cite{Phong1998Illumination}, for spatially consistent normal directions, we employ barycentric interpolation and interpolate the vertex normals via barycentric coordinates ($\alpha+\beta+\gamma=1,\,\alpha\bm{v}_i+\beta\bm{v}_j+\gamma\bm{v}_k=\bm{p}$):
\begin{equation}
\mathbf{n}_p = \frac{\alpha \bm{n}_{v_i} + \beta \bm{n}_{v_j} + \gamma \bm{n}_{v_k}}{\|\alpha \bm{n}_{v_i} + \beta \bm{n}_{v_j} + \gamma \bm{n}_{v_k}\|_2}\,,
\end{equation}
where the vertex normals are computed by averaging the adjacent face normals. This strategy ensures a smooth normal field across the surface, resulting in improved view synthesis quality, as evidenced by the ablation studies (\tab~\ref{tab:Ablation}).

\begin{algorithm}[t]
\caption{Recursive Subdivision Tracing}
\label{alg:tracing}
\begin{algorithmic}[1]
\Statex \textbf{Input:}
  An initial mesh proxy $\mathcal{M}_0=(\mathcal{V}_0,\mathcal{F}_0)$, the ray queries $\bm{r}$, the height field $\mathcal{H}$.
\Statex \textbf{Output:}
  The ray depth $t$ with intersection normal $\mathbf{n}$.
\State \codecomment{Start with a coarse mesh for lightweight computing.}
\State $\mathcal{V}_0 \leftarrow \text{QueryHeightField}(\mathcal{H},\mathcal{V}_0)$
\State $\mathcal{B} \leftarrow \text{BuildBVH}(\mathcal{V}_0,\mathcal{F}_0)$
\State \codecomment{Identify the intersected triangle and produce a mask.}
\State $M_\mathrm{hit} \leftarrow \text{GlobalIntersetionCompute}(\mathcal{B},\bm{r})$
\For{$i = 1 \text{ to } K$}
    \State \codecomment{Remove unhit vertices to reduce MLP overhead.}
    \State $(\mathcal{V}_{i-1}',\mathcal{F}_{i-1}') \leftarrow \text{RemoveUnhit}(\mathcal{V}_{i-1},\mathcal{F}_{i-1},{M}_\mathrm{hit})$
    \State \codecomment{Each triangle is subdivided into four. }
    \State $(\mathcal{V}_i,\mathcal{F}_i) \leftarrow \text{SubdivideTriangles}(\mathcal{V}_{i-1}',\mathcal{F}_{i-1}')$
    \State \codecomment{Refine the positions of newly added vertices. }
    \State $\mathcal{V}_i \leftarrow \text{QueryHeightField}(\mathcal{H},\mathcal{V}_i)$
    \State \codecomment{Only the four sub-triangles are checked. }
    \State $t,\mathbf{n},{M}_\mathrm{hit} \leftarrow \text{LocalIntersectionCompute}(\mathcal{V}_i,\bm{r})$
\EndFor
\State \Return $t,\mathbf{n}$
\end{algorithmic}
\end{algorithm}

\subsection{Refraction-Aware Gaussian Ray Tracing}
\label{sec:method:grt}
Our ray tracing process integrates the hierarchical surface model with the 3DGS framework. Similar to other ray-based methods, we begin by casting a ray from each pixel.

\paragraph{Forward}
Each ray first intersects the refractive surface, yielding an intersection point $\bm{p}$ and normal $\mathbf{n}$ from the method described in \sec~\ref{sec:method:surface}. Using these, the refracted ray direction $\mathbf{\omega_o}$ is computed via Snell’s law (Eq.~\ref{eq:snell:vector}). This new ray, originating from $\bm{p}$ with direction $\mathbf{\omega_o}$, then traverses the underlying scene. It intersects a series of Gaussian primitives, accumulating color and opacity as in standard 3DGS~\cite{kerbl3Dgaussians}. These accumulated colors are subsequently used for loss computation noted in \eq~\ref{eq:total_loss}.

\paragraph{Backward}
During backpropagation, the gradients are twofold. First, color gradients are propagated along each ray to all Gaussian primitives that contributed to the pixel, optimizing the underlying scene's parameters (position, opacity, SH, etc.).
Different from 3DGRT~\cite{loccoz20243dgrt}, our method additionally computes the gradients of the rendered color with respect to the refracted ray's origin $\bm{p}$ and direction $\mathbf{\omega_o}$. These gradients are then propagated back to the parameters of our hierarchical surface model (i.e., the weights of the MLP $\mathcal{H}(x,y)$). This joint optimization allows the framework to simultaneously reconstruct the underlying scene and the refractive surface geometry that best explains the observed images. We detail the gradient computation in the supplementary material.

\subsection{Implementation Details}
\label{sec:method:impl}
We build our method upon the open-source 3DGRUT~\cite{loccoz20243dgrt, wu20253dgut} codebase, adapting it for refractive ray tracing. We implement our water height map in PyTorch and accelerate the ray–mesh intersection and subdivision using NVIDIA OptiX and custom CUDA kernels. We train our model for 15k steps across all experimental scenes. More details are in the supplementary material.

\paragraph{Loss Functions}
Consistent with prior works~\cite{kerbl3Dgaussians,loccoz20243dgrt}, we conduct optimization under multi-view image supervision, employing a combination of $\mathcal{L}_{\text{L1}} = \| \mathbf{C} - \mathbf{C}_{gt} \|_1$ loss and SSIM loss $\mathcal{L}_{\text{SSIM}} = 1 - \text{SSIM}(\mathbf{C}, \mathbf{C}_{gt})$ as the primary supervision signal between the rendered image $\mathbf{C}$ and the ground-truth image $\mathbf{C}_{gt}$. During training, we observe that sparse‑view regions produce high‑opacity floaters, which occlude subsequent Gaussians and cause gradient vanishing as analyzed in~\cite{wang2025stablegs}. To mitigate this, we additionally introduce an opacity loss $\mathcal{L}_{\alpha}$ that penalizes high opacity values across all $N$ Gaussians:
\begin{equation}
\mathcal{L}_{\alpha} = \frac{1}{N} \sum_{i=1}^N \alpha_i^2\,.
\end{equation}
As illustrated in \fig~\ref{fig:opacity_loss_effect}, this simple regularization effectively encourages transparency and stabilizes optimization, especially in sparsely-viewed regions. Finally, our total loss is a weighted sum of these components, where we set $\lambda_1=0.8$, $\lambda_2=0.2$ and $\lambda_3=0.007$:
\begin{equation}
\mathcal{L}_{\text{total}} = \lambda_1 \mathcal{L}_{\text{L1}} + \lambda_2 \mathcal{L}_{\text{SSIM}} + \lambda_3 \mathcal{L}_{\alpha}\,.
\label{eq:total_loss}
\end{equation}

\begin{comment}

\end{comment}

%% file: sec/4_experiments.tex
\section{Experiments}
\label{sec:experiments}

\subsection{Experimental Setup}

\paragraph{Datasets} 
To rigorously validate our method, we conduct evaluation on datasets where multi-view observation of an underwater scene is captured from above the non-planar water surface. Specifically, we select two datasets covering both synthetic and real-world scenarios:

\noindent\emph{NeRFrac Dataset.} Our experiments cover the entire NeRFrac~\cite{zhan2023nerfrac} benchmark, which includes \textbf{4 real-world scenes} and \textbf{3 synthetic scenes}, each with 8 training views and 1 test view looking vertically down onto the water surface. The image resolution of the real dataset is $384 \times 512$, while that of the synthetic dataset is $392 \times 392$.

\noindent\emph{\oursdataset~Dataset.} To further evaluate our method under stronger refractive effects, we construct the \oursdataset~dataset. Compared with NeRFrac’s narrow-baseline setup, \oursdataset~features scenes captured with larger camera angular coverage, where rays hit the water surface at higher incidence angles, amplifying refraction and making reconstruction more challenging. Each scene consists of 24 training views in a surround-view configuration and 6 test views. In addition, \oursdataset~offers ground‑truth water‑surface geometry, which is not publicly accessible in the NeRFrac dataset. Further dataset details and rendering settings are provided in the supplementary material.

\paragraph{Baselines}
We benchmark our method against both refractive reconstruction and general scene reconstruction approaches. For refractive reconstruction, we include WSGS~\cite{yoon2026wsgs}, which is constrained to planar refractive water surfaces, and NeRFrac~\cite{zhan2023nerfrac}, which represents the state-of-the-art in modeling arbitrary, non-planar refractive fluid surface. For general radiance field reconstruction, we include representative methods across different paradigms: Mip-NeRF~\cite{barron2021mipnerf}, Tensorf~\cite{Chen2022tensorf}, Plenoxels~\cite{yu2022plenoxels}, 3DGS~\cite{kerbl3Dgaussians}, and 3DGRT~\cite{loccoz20243dgrt}. This selection covers both volumetric and point-based representations, providing a comprehensive baseline for evaluating our approach. All baselines were trained with their default settings. For fairness, we also report 3DGS and 3DGRT results trained with fewer iterations (under the same setting as ours, 15k instead of default 30k), provided in the supplementary material.

\paragraph{Metrics}
We evaluate the performance from two aspects: \emph{novel-view synthesis quality} and \emph{refractive surface reconstruction accuracy}.  For novel view synthesis quality, we report PSNR, SSIM~\cite{wang2004ssim}, and LPIPS~\cite{zhang2018lpips} scores. For refractive surface reconstruction accuracy, we report the root mean square error (RMSE) of the minimum distance between the intersection point of each ray from the test views and the ground-truth water surface. All experiments were conducted on a single RTX 4090 GPU.

\input{sec/tab/quantitative}

\subsection{Experimental Results}

\paragraph{Quantitative Results}
\tab~\ref{tab:Main_Results} presents the quantitative results. 
On the NeRFrac dataset, our method significantly outperforms all existing approaches. Compared to the previous state-of-the-art, it achieves a PSNR increase of over 2+ on real scenes and 1.4+ on synthetic scenes. While WSGS formulates an physics-based refraction model, its performance inevitably degrades on these non-planar datasets due to its strong reliance on a perfectly planar water surface.
On the more challenging \oursdataset~dataset, all prior methods, including NeRFrac, struggle to handle severe refractive distortions, while our approach consistently delivers high-fidelity novel views and precise refractive-surface reconstruction.

Benefiting from our efficient refractive‑surface intersection and the rendering efficiency of Gaussian representations, our method trains substantially faster than NeRFrac while maintaining real‑time rendering.
It also converges more quickly than 3DGRT, as the absence of refraction modeling in 3DGRT leads to inaccurate scene learning and excessive cloning and splitting of Gaussians, which slow down optimization. Its longer default training schedule further increases runtime. 

\input{sec/tab/qualitative}

\paragraph{Qualitative Results}
\fig~\ref{fig:results} presents qualitative comparisons demonstrating the superiority of our method. On the NeRFrac dataset, both our method and NeRFrac achieve comparable reconstruction quality, but our approach better preserves fine textures and high-frequency details, whereas NeRFrac produces noticeably smoother results. On the more challenging \oursdataset~dataset, where refractionve distortions are severe, existing methods fail to reconstruct consistent views. In contrast, our method achieves photorealistic rendering across all test scenes.

\input{sec/tab/abulation}

\paragraph{Ablation Studies}
We perform ablation studies to analyze the contribution of each module to reconstruction quality. Specifically, we conduct four independent experiments: replacing our \textbf{(a) water height map} (\ref{sec:method:height}) with NeRFrac’s refractive field, disabling \textbf{(b) normal smoothing} (\ref{sec:method:normal}), disabling \textbf{(c) recursive intersection} (\ref{sec:method:subdivision}), and disabling the \textbf{(d) opacity loss} (\ref{sec:method:impl}). Results are summarized in \tab~\ref{tab:Ablation}, all conducted on the NeRFrac real and synthetic dataset.
Our refractive-surface representation provides significant improvements in both accuracy and efficiency compared to the refraction-field representation. The normal-smoothing strategy enhances rendering quality by promoting smoother surface shading, with only minor computational overhead. Recursive subdivision slightly reduces rendering quality, as a few rays may miss micro‑triangles, but it shortens training time by nearly 50\%, offering a favorable efficiency–accuracy trade‑off. Finally, the opacity loss effectively suppresses local floaters, resulting in noticeably improved visual quality.

\subsection{Applications}

\paragraph{Water Removal}
\input{sec/tab/water_removal}

Similar to NeRFrac, our approach is also capable of removing the refractive water-surface effect. After training, this can be achieved by disabling the refraction computation that bends ray trajectories during rendering. We additionally evaluate the water removal image quality of our proposed method and NeRFrac on \emph{\oursdataset~dataset}, as shown in \tab~\ref{tab:Water_Removal_half}. Benefiting from our high-precision water surface representation, we achieve high-quality water removal results, which further enable us to extract the geometry of underwater objects from the dewatered images using geometry extraction methods such as PGSR~\cite{chen2024pgsr}, as shown in~\fig~\ref{fig:application}. 

\paragraph{Water Surface Editing}
Compared with NeRFrac, an additional advantage of our explicit water surface representation is its support for surface editing. Once a scene has been trained, the water surface can be replaced with any arbitrary mesh without the need for retraining. \fig~\ref{fig:application} illustrates this capability, where we substitute the original surface with a triangular wave surface.

%% file: sec/tab/quantitative.tex
\begin{table*}[t]
  \centering
  \caption{\textbf{Quantitative experimental results.} Our method achieves state-of-the-art novel view synthesis quality and water surface reconstruction accuracy, while maintaining fast training speed and real-time rendering frame rates. ($\dagger$ indicates methods specifically designed for refractive scenes.)}
  \label{tab:Main_Results}

  \setlength{\tabcolsep}{8pt}

  \resizebox{\linewidth}{!}{%
  \begin{tabular}{ll | cccccc}
    \toprule
    Dataset & Method & PSNR$\uparrow$ & SSIM$\uparrow$ & LPIPS$\downarrow$ & Train$\downarrow$ & FPS$\uparrow$ & RMSE(cm)$\downarrow$ \\
    \midrule
    
    % === Dataset 1: NeRFrac Real ===
    \multirow{7}{*}{NeRFrac Real} 
    & Mip-NeRF  & 10.385 & 0.221 & 0.916 & $\sim$15 h & $<$ 1 & - \\
    & TensoRF   & 22.973 & 0.809 & \cellcolor{thirdscore!80}0.139 & 11 min & $<$ 1 & - \\
    & Plenoxels & 14.867 & 0.389 & 0.631 & \cellcolor{secondscore!80}9 min & 61 & - \\
    & 3DGS      & 27.711 & \cellcolor{thirdscore!80}0.864 & \cellcolor{secondscore!80}0.130 & \cellcolor{firstscore!80}3 min & \cellcolor{firstscore!80}857 & - \\
    & 3DGRT     & \cellcolor{thirdscore!80}27.732 & 0.842 & 0.164 & 16 min & \cellcolor{thirdscore!80}157 & - \\
    & WSGS$^\dagger$  & 22.449 & 0.679 &  0.279 & 106 min & 129 & - \\
    & NeRFrac$^\dagger$   & \cellcolor{secondscore!80}28.146 & \cellcolor{secondscore!80}0.876 & 0.167 & 149 min & $<$ 1 & - \\
    & \textbf{Ours$^\dagger$} & \cellcolor{firstscore!80}30.671 & \cellcolor{firstscore!80}0.913 & \cellcolor{firstscore!80}0.129 & \cellcolor{thirdscore!80}10 min & \cellcolor{secondscore!80}242 & - \\
    
    \midrule
    
    % === Dataset 2: NeRFrac Synthetic ===
    \multirow{7}{*}{NeRFrac Synthetic} 
    & Mip-NeRF  & 12.414 & 0.382 & 0.870 & $\sim$15 h & $<$ 1 & - \\
    & TensoRF   & 16.223 & 0.532 & 0.601 & \cellcolor{secondscore!80}11 min & $<$ 1 & - \\
    & Plenoxels & 13.334 & 0.407 & 0.806 & \cellcolor{thirdscore!80}12 min & 72 & - \\
    & 3DGS      & \cellcolor{thirdscore!80}22.186 & \cellcolor{thirdscore!80}0.730 & 0.300 & \cellcolor{firstscore!80}3 min & \cellcolor{firstscore!80}704 & - \\
    & 3DGRT     & 21.929 & 0.689 & 0.363 & 27 min & 89 & - \\
    & WSGS$^\dagger$  & 20.868 &  0.726 &  \cellcolor{thirdscore!80}0.271 &  107 min & \cellcolor{thirdscore!80}153 & - \\
    & NeRFrac$^\dagger$   & \cellcolor{secondscore!80}34.381 & \cellcolor{secondscore!80}0.944 & \cellcolor{secondscore!80}0.149 & 77 min & $<$ 1 & - \\
    & \textbf{Ours$^\dagger$} & \cellcolor{firstscore!80}35.816 & \cellcolor{firstscore!80}0.952 & \cellcolor{firstscore!80}0.136 & \cellcolor{thirdscore!80}12 min & \cellcolor{secondscore!80}203 & - \\

    \midrule
    
    % === Dataset 3: Your Dataset (\oursdataset) ===
    \multirow{7}{*}{\oursdataset} 
    & Mip-NeRF  & 13.874 & \cellcolor{secondscore!80}0.479 & 0.808 & $\sim$15 h & $<$ 1 & - \\
    & TensoRF   & 14.125 & 0.362 & 0.658 & 13 min & $<$ 1 & - \\
    & Plenoxels & 13.983 & 0.383 & 0.790 & \cellcolor{thirdscore!80}12 min & 23 & - \\
    & 3DGS      & 17.482 & \cellcolor{thirdscore!80}0.428 & \cellcolor{thirdscore!80}0.486 & \cellcolor{firstscore!80}5 min & \cellcolor{firstscore!80}593 & - \\
    & 3DGRT     & \cellcolor{secondscore!80}18.019 & 0.424 & 0.545 & 14 min & \cellcolor{secondscore!80}182 & - \\
    & WSGS$^\dagger$  & \cellcolor{thirdscore!80}17.915 &  0.424 & \cellcolor{secondscore!80}0.484 & 105 min & \cellcolor{thirdscore!80}124 & - \\
    & NeRFrac$^\dagger$   & 17.837 & 0.405 & 0.624 & 164 min & $<$ 1 & \cellcolor{secondscore!80}3.655 \\
    & \textbf{Ours$^\dagger$} & \cellcolor{firstscore!80}30.224 & \cellcolor{firstscore!80}0.933 & \cellcolor{firstscore!80}0.098 & \cellcolor{secondscore!80}11 min & \cellcolor{thirdscore!80}124 & \cellcolor{firstscore!80}0.115 \\

    \bottomrule
  \end{tabular}%
  }
  \vspace*{-2mm}
\end{table*}

%% file: sec/tab/qualitative.tex
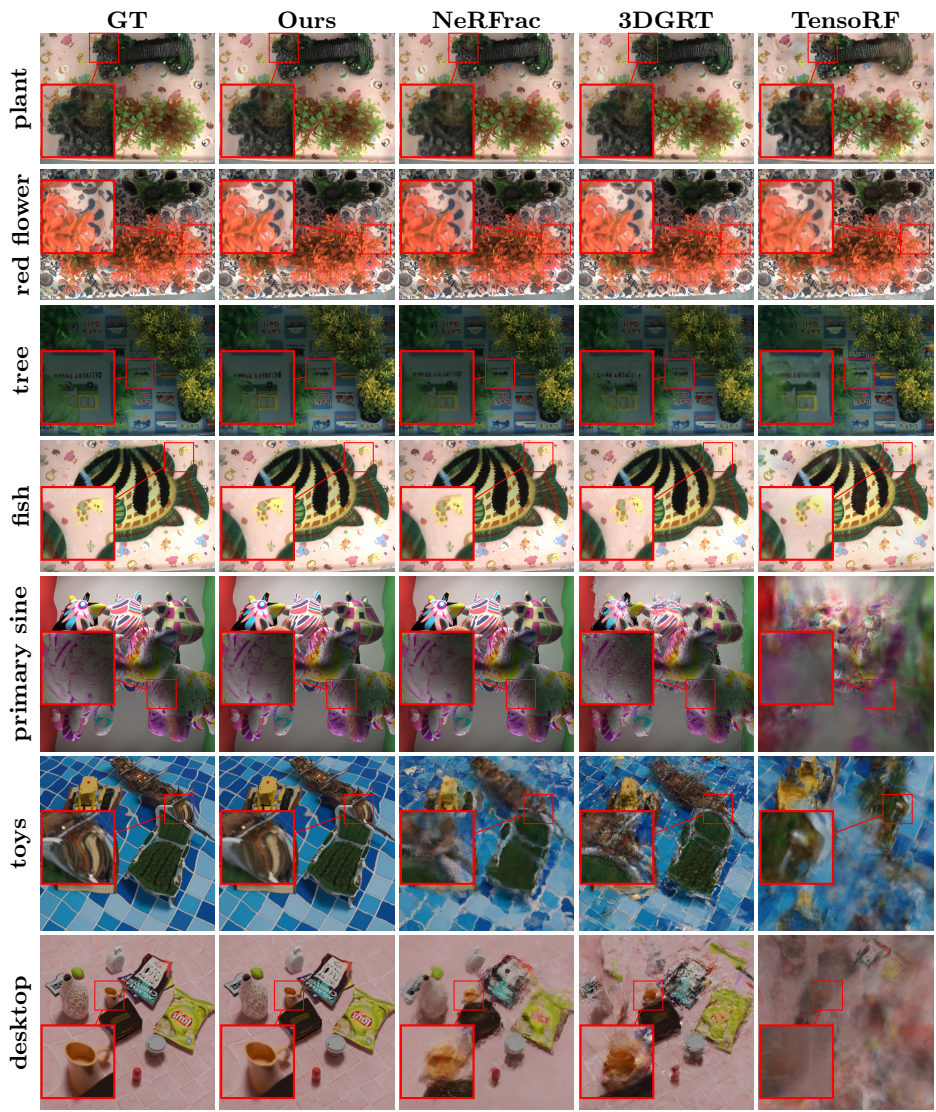
\begin{figure*}
    \centering
    \setlength{\tabcolsep}{1pt}
    \setlength{\imagewidth}{0.19\textwidth}
    \renewcommand{\arraystretch}{0.6}
    \newcommand{\formattedgraphicsfirst}[1]{%
      \begin{tikzpicture}[spy using outlines={rectangle, magnification=2.5, connect spies}]
        \node[anchor=south west, inner sep=0] at (0,0){\includegraphics[width=\imagewidth]{#1}};
        \spy [red,size=27.5pt] on (.37\imagewidth,.67\imagewidth) in node at (.22\imagewidth,.25\imagewidth);
        \end{tikzpicture}%
    }
    \newcommand{\formattedgraphicssecond}[1]{%
      \begin{tikzpicture}[spy using outlines={rectangle, magnification=2.5, connect spies}]
        \node[anchor=south west, inner sep=0] at (0,0){\includegraphics[width=\imagewidth]{#1}};
        \spy [red,size=27.5pt] on (.7\imagewidth,.33\imagewidth) in node at (.22\imagewidth,.48\imagewidth);
        \end{tikzpicture}%
    }
    \newcommand{\formattedgraphicsthird}[1]{%
      \begin{tikzpicture}[spy using outlines={rectangle, magnification=2.5, connect spies}]
        \node[anchor=south west, inner sep=0] at (0,0){\includegraphics[width=\imagewidth]{#1}};
        \spy [red,size=27.5pt] on (.8\imagewidth,.7\imagewidth) in node at (.22\imagewidth,.48\imagewidth);
        \end{tikzpicture}%
    }
    \newcommand{\formattedgraphicsfourth}[1]{%
      \begin{tikzpicture}[spy using outlines={rectangle, magnification=2.5, connect spies}]
        \node[anchor=south west, inner sep=0] at (0,0){\includegraphics[width=\imagewidth]{#1}};
        \spy [red,size=27.5pt] on (.9\imagewidth,.35\imagewidth) in node at (.22\imagewidth,.48\imagewidth);
        \end{tikzpicture}%
    }
    \newcommand{\formattedgraphicsfifth}[1]{%
      \begin{tikzpicture}[spy using outlines={rectangle, magnification=2.5, connect spies}]
        \node[anchor=south west, inner sep=0] at (0,0){\includegraphics[width=\imagewidth]{#1}};
        \spy [red,size=27.5pt] on (.4\imagewidth,.66\imagewidth) in node at (.22\imagewidth,.28\imagewidth);
        \end{tikzpicture}%
    }
    \newcommand{\formattedgraphicstree}[1]{%
      \begin{tikzpicture}[spy using outlines={rectangle, magnification=2.5, connect spies}]
        \node[anchor=south west, inner sep=0] at (0,0){\includegraphics[width=\imagewidth]{#1}};
        \spy [red,size=27.5pt] on (.58\imagewidth,.36\imagewidth) in node at (.22\imagewidth,.28\imagewidth);
        \end{tikzpicture}%
    }
    \newcommand{\formattedgraphicsfish}[1]{%
      \begin{tikzpicture}[spy using outlines={rectangle, magnification=2.5, connect spies}]
        \node[anchor=south west, inner sep=0] at (0,0){\includegraphics[width=\imagewidth]{#1}};
        \spy [red,size=27.5pt] on (.8\imagewidth,.66\imagewidth) in node at (.22\imagewidth,.28\imagewidth);
        \end{tikzpicture}%
    }
    \begin{tabular}{m{0.38cm}<{\centering}m{\imagewidth}<{\centering}m{\imagewidth}<{\centering}m{\imagewidth}<{\centering}m{\imagewidth}<{\centering}m{\imagewidth}<{\centering}}
        & \textbf{GT} & \textbf{Ours} & \textbf{NeRFrac} & \textbf{3DGRT} & \textbf{TensoRF} \\
        \rotatebox{90}{\textbf{plant}} &
        \formattedgraphicsfirst{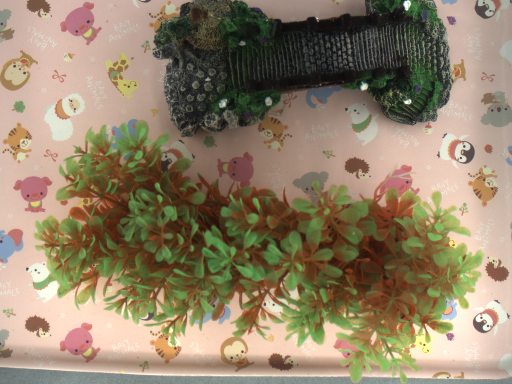} & 
        \formattedgraphicsfirst{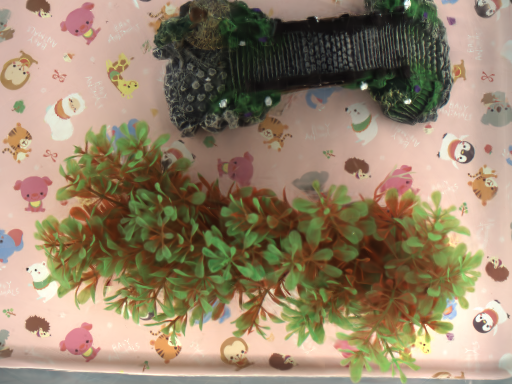} & 
        \formattedgraphicsfirst{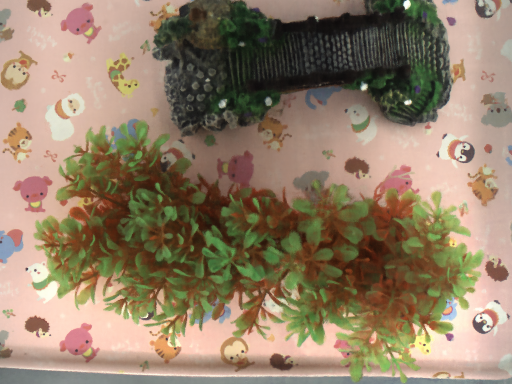} &
        \formattedgraphicsfirst{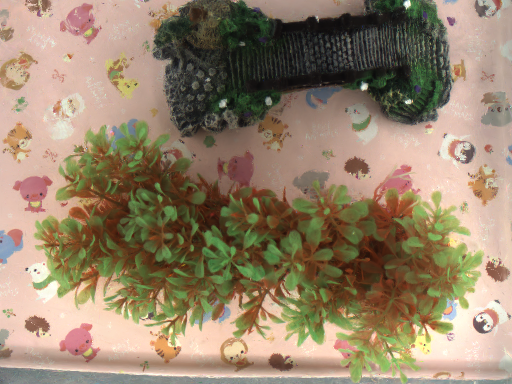} & 
        \formattedgraphicsfirst{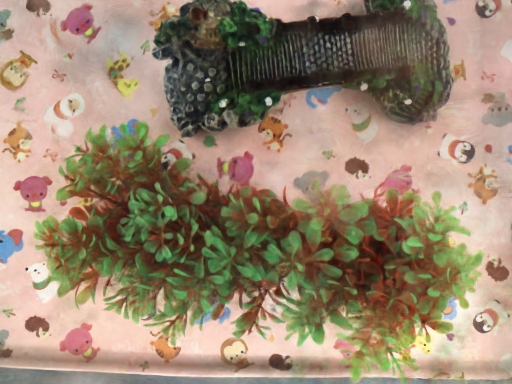} \\
        \rotatebox{90}{\textbf{red flower}} &
        \formattedgraphicsfourth{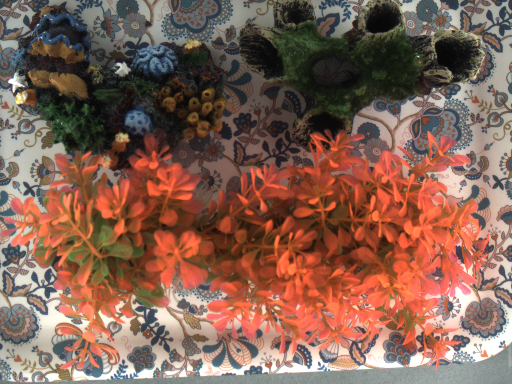} & 
        \formattedgraphicsfourth{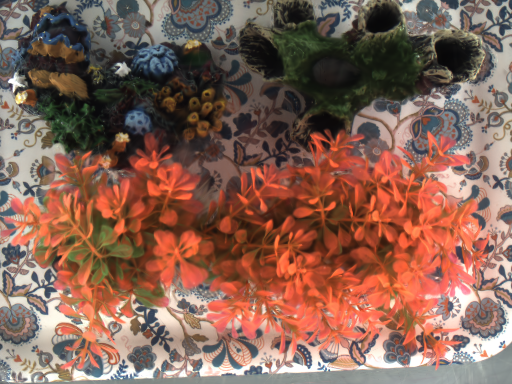} & 
        \formattedgraphicsfourth{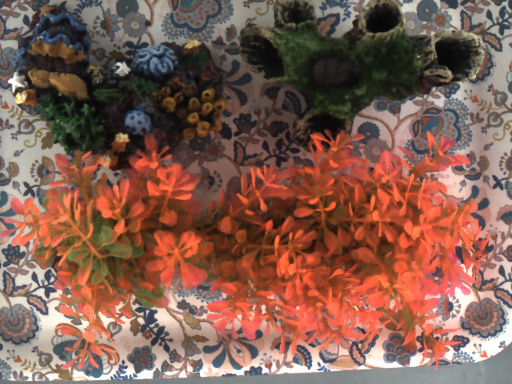} &
        \formattedgraphicsfourth{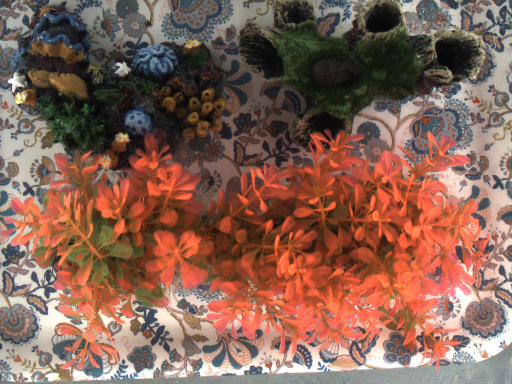} & 
        \formattedgraphicsfourth{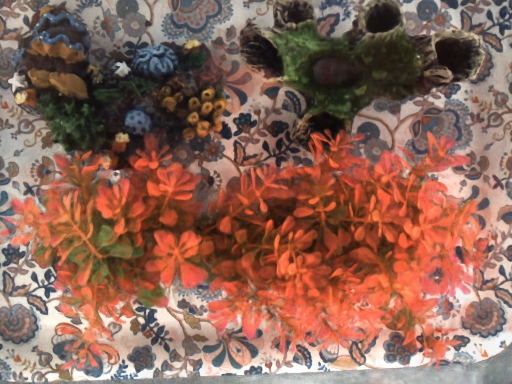} \\

        \rotatebox{90}{\textbf{tree}} &
        \formattedgraphicstree{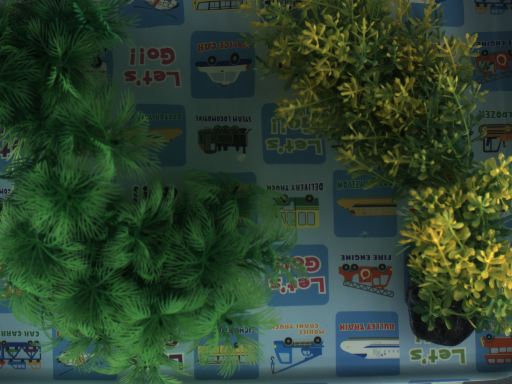} & 
        \formattedgraphicstree{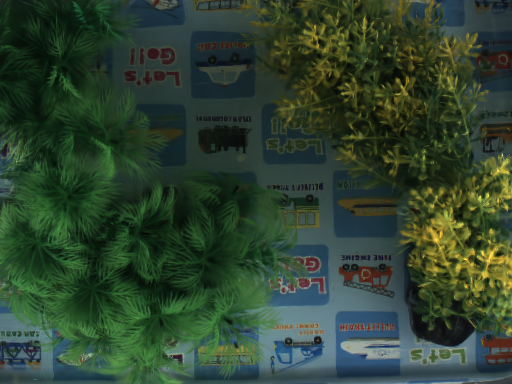} & 
        \formattedgraphicstree{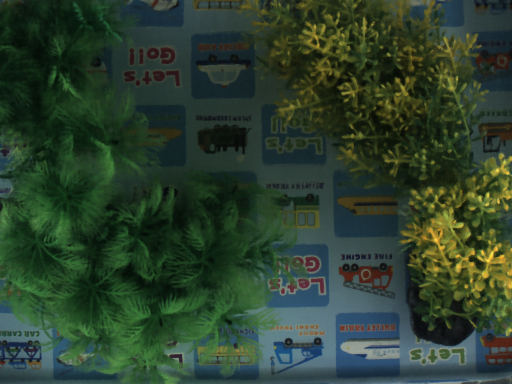} &
        \formattedgraphicstree{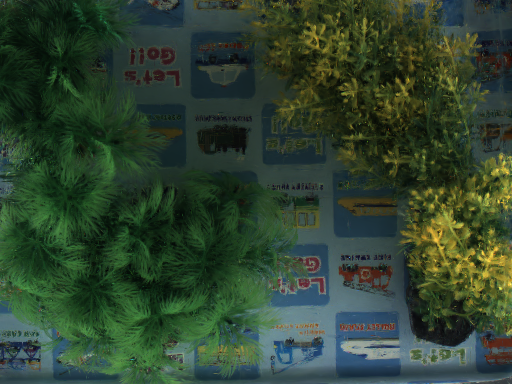} & 
        \formattedgraphicstree{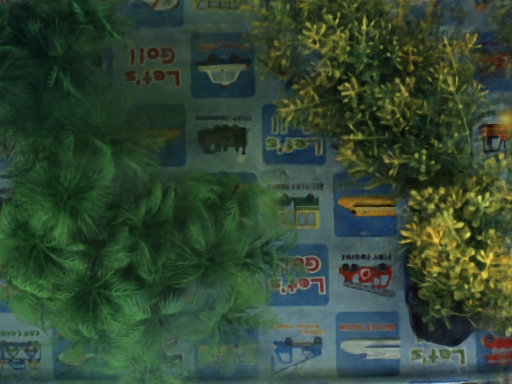} \\

        \rotatebox{90}{\textbf{fish}} &
        \formattedgraphicsfish{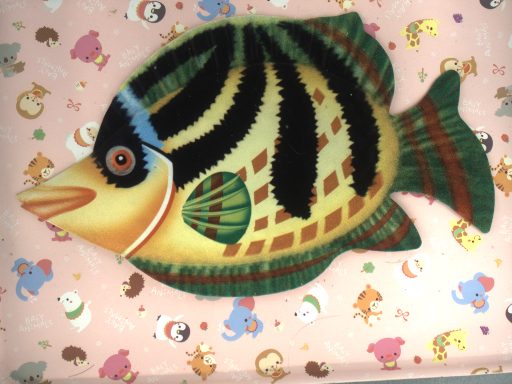} & 
        \formattedgraphicsfish{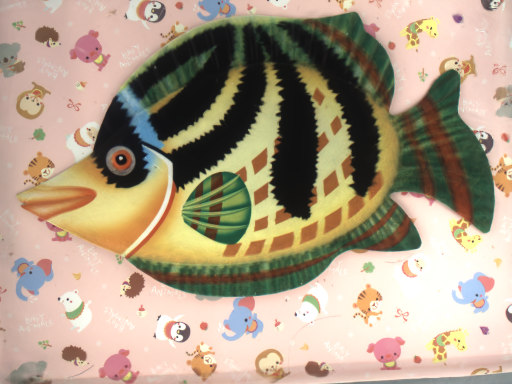} & 
        \formattedgraphicsfish{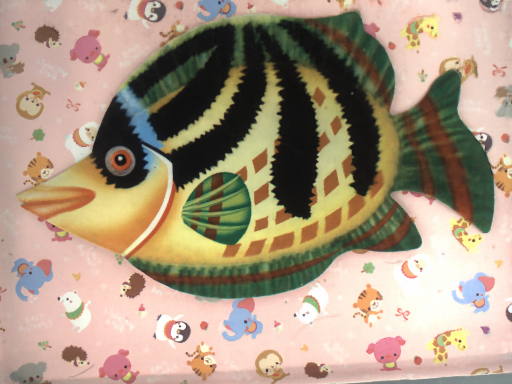} &
        \formattedgraphicsfish{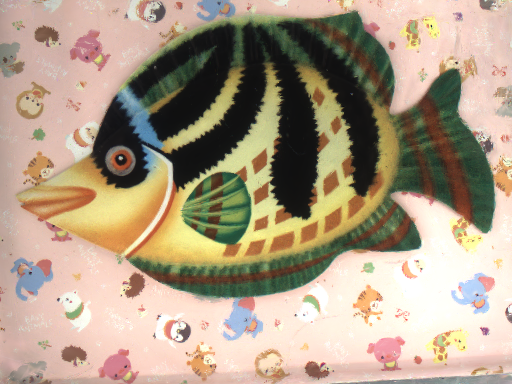} & 
        \formattedgraphicsfish{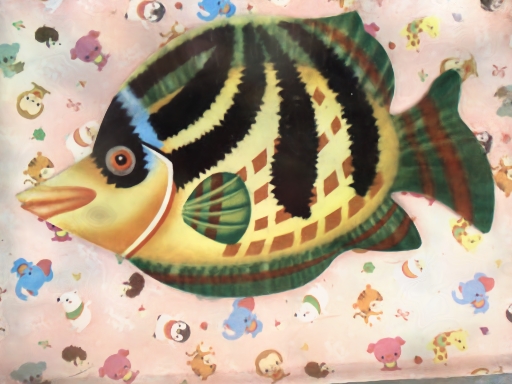} \\
        
        \rotatebox{90}{\textbf{primary sine}} &
        \formattedgraphicssecond{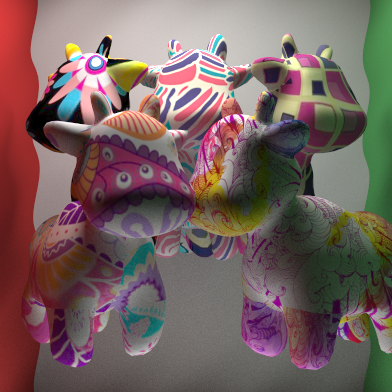} & 
        \formattedgraphicssecond{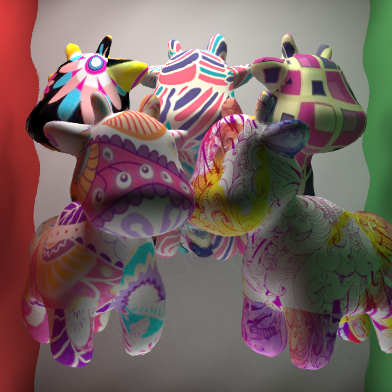} & 
        \formattedgraphicssecond{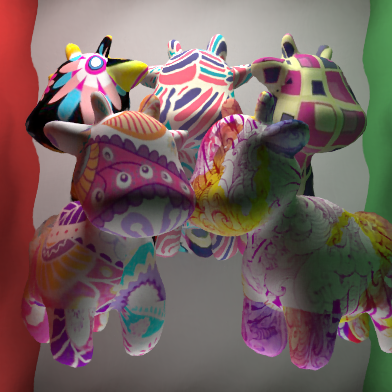} &
        \formattedgraphicssecond{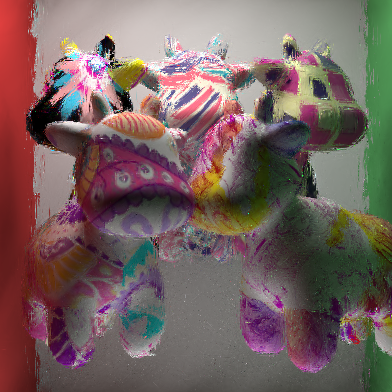} & 
        \formattedgraphicssecond{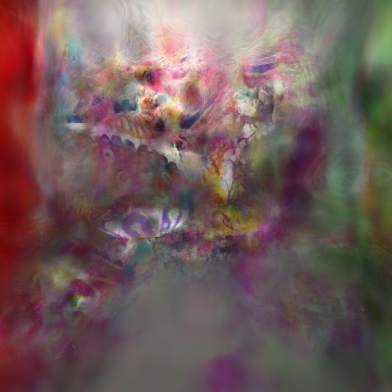} \\
        \rotatebox{90}{\textbf{toys}} & 
        \formattedgraphicsthird{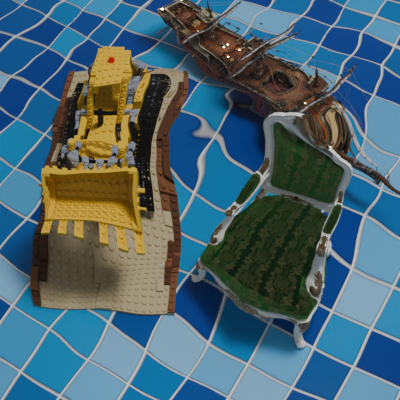} & 
        \formattedgraphicsthird{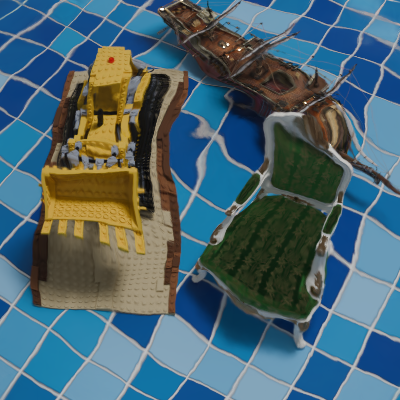} & 
        \formattedgraphicsthird{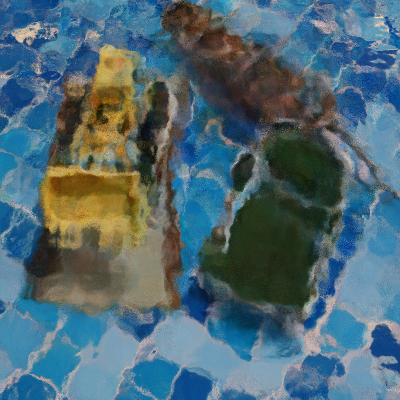} &
        \formattedgraphicsthird{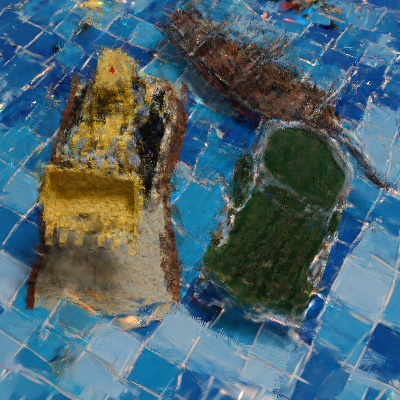} & 
        \formattedgraphicsthird{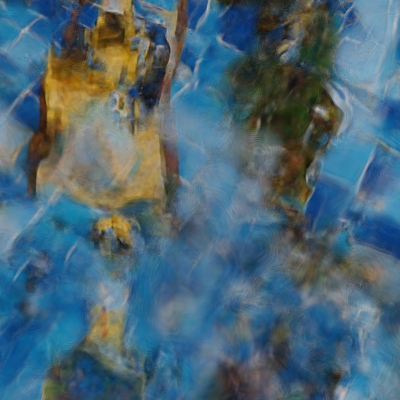} \\
        \rotatebox{90}{\textbf{desktop}} & 
        \formattedgraphicsfifth{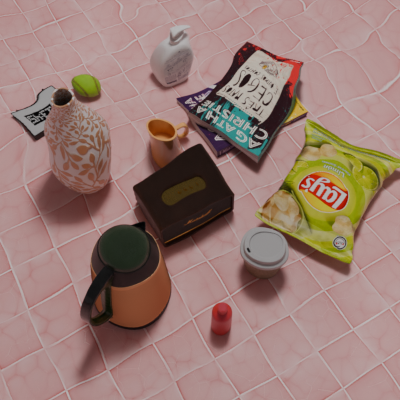} & 
        \formattedgraphicsfifth{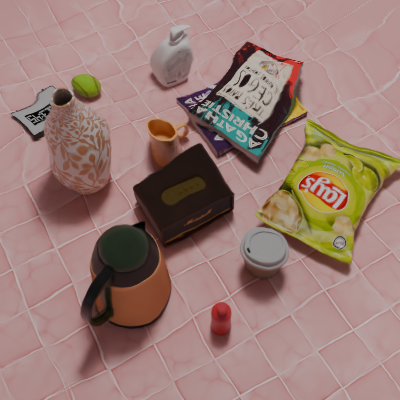} & 
        \formattedgraphicsfifth{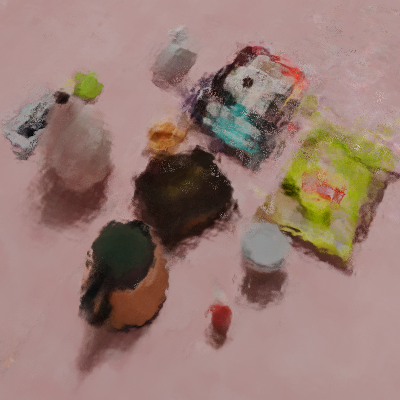} &
        \formattedgraphicsfifth{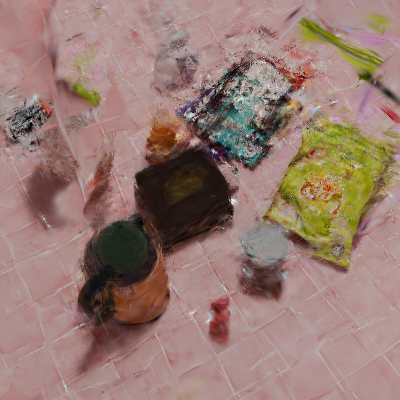} & 
        \formattedgraphicsfifth{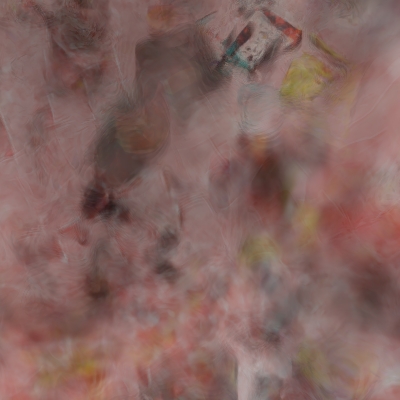} \\
    \end{tabular}
    \vspace*{-3mm}
    \caption{\label{fig:results} 
    \textbf{Qualitative comparison with existing methods.} On the NeRFrac dataset, our method preserves fine textures and high-frequency details better than NeRFrac. On the more challenging \oursdataset~dataset—featuring wider viewpoint coverage and stronger refractive effects—our method consistently produces high-fidelity renderings, whereas existing methods fail to learn accurate scene representations under such conditions, leading to noticeable artifacts and incorrect novel-view synthesis.}
    \vspace*{-3mm}
\end{figure*}

%% file: sec/tab/abulation.tex
\begin{table}[t]
  \centering
  \caption{\textbf{Ablation study results.} The ablations show that the components of our method collectively provide a favorable balance between reconstruction quality and computational efficiency.}
  \vspace{-1mm}
  \label{tab:Ablation}

  \setlength{\tabcolsep}{12pt}

  \resizebox{\linewidth}{!}{%
  \begin{tabular}{l | ccccc }
    \toprule
    Method\textbar Metric & PSNR$\uparrow$ & SSIM$\uparrow$ & LPIPS$\downarrow$ & Train$\downarrow$ & FPS$\uparrow$   \\
    \midrule
    (a) w/o Water Height Map & 27.782 & 0.869 & 0.168 & 22 min & 57 \\
    (b) w/o Normal Smoothing & \cellcolor{thirdscore!80}32.643 & \cellcolor{thirdscore!80}0.928 & 0.133 & \cellcolor{firstscore!80}11 min & \cellcolor{firstscore!80}232 \\
    (c) w/o Recursive Intersection & \cellcolor{firstscore!80}32.966 & \cellcolor{firstscore!80}0.929 & \cellcolor{secondscore!80}0.132 & 20 min & 218 \\
    (d) w/o Opacity Loss & 31.927 & 0.920 & \cellcolor{firstscore!80}0.125 & \cellcolor{secondscore!80}12 min & \cellcolor{secondscore!80}224 \\
    Full Model & \cellcolor{secondscore!80}32.876 & \cellcolor{firstscore!80}0.929 & \cellcolor{secondscore!80}0.132 & \cellcolor{firstscore!80}11 min & \cellcolor{thirdscore!80}223\\
    \bottomrule
  \end{tabular}%
  }
  \vspace{-1mm}
\end{table}

%% file: sec/tab/water_removal.tex
\begin{wraptable}{r}{0.5\textwidth}
    \centering 
    \vspace{-12mm}
    \caption{\textbf{Metrics of water removal quality.} Our method yields more reliable results compared with NeRFrac.}
    \label{tab:Water_Removal_half}

    \setlength{\tabcolsep}{5pt} 
    \resizebox{\linewidth}{!}{%
    \begin{tabular}{l | ccc }
        \toprule
        Method\textbar Metric & PSNR$\uparrow$ & SSIM$\uparrow$ & LPIPS$\downarrow$    \\
        \midrule
        NeRFrac & 14.950 & 0.380 & 0.563 \\
        Ours & \textbf{23.151} & \textbf{0.869} & \textbf{0.110} \\
        \bottomrule
    \end{tabular}%
    }
    \vspace{-6mm}
\end{wraptable}

%% file: sec/5_conclusion.tex
\begin{figure*}[t]
	\centering
	\includegraphics[width=0.9\linewidth]{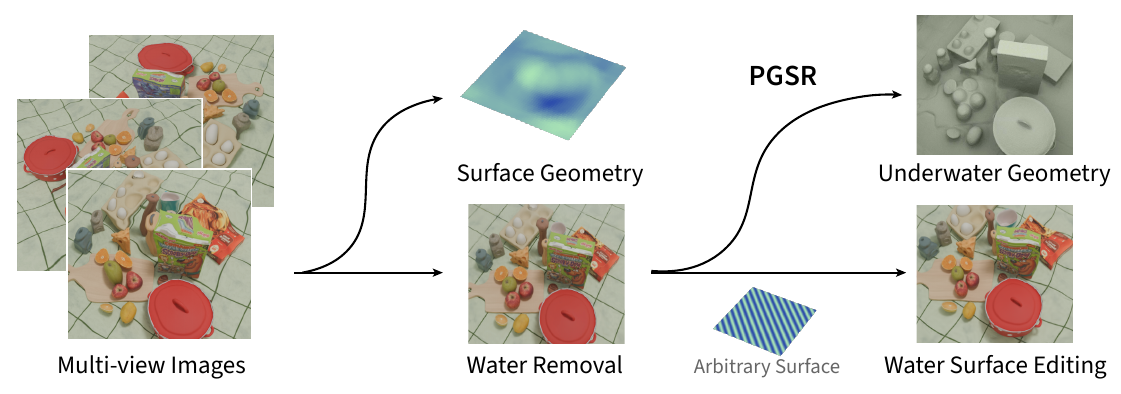}
    \caption{
    \label{fig:application} 
    \textbf{Water removal and surface editting} Benefiting from our physically-based modeling approach, our method naturally decouples the water surface geometry from underwater objects. This capability enables the removal of surface refraction, as well as water surface editing using arbitrary meshes. Crucially, our precise refraction modeling allows these refraction-removed views to serve directly as inputs for methods like PGSR to achieve complete geometric reconstruction of the underlying scene.
    }
\end{figure*}

\section{Conclusions and Limitations}
\label{sec:conclusion}

\paragraph{Conclusions} 
We present {\name}, a framework for reconstructing scenes through refractive surfaces with 3D Gaussian ray tracing. By introducing an explicit, differentiable refractive‑surface model and refraction-aware Gaussian ray tracing, our method enables accurate and efficient modeling of both the interface and the scene. Experiments on both real and synthetic datasets show that {\name} outperforms prior methods in rendering quality, reconstruction accuracy, and training efficiency. The explicit surface also supports flexible applications such as water removal and surface editing.

\paragraph{Limitations}
While our approach demonstrates encouraging performance in handling refractive water surfaces, it inherently charts several promising avenues for future research.
First, our formulation prioritizes refraction-induced geometric distortions; accounting for features like reflection and scattering will further enhance the method's application in a wider range of real-world scenarios.
Second, our current framework is tailored for static scenes, serving as a foundational stepping stone toward the more challenging modeling of dynamic underwater environments. 
Third, we adopt a single-value height field assumption for the water surface, which leaves rare, highly turbulent phenomena like curling or breaking waves as an open invitation for more complex geometric representations. 
Fourth, although 3DGS guarantees high-fidelity view synthesis, exploring the integration of 2DGS~\cite{Huang2DGS2024} offers a compelling path forward to balance rendering quality with precise surface geometry for exact reconstruction tasks. 
We provide a more expansive discussion of these forward-looking directions in the supplementary material.

\section{Acknowledgment}
This work was supported by the National Natural Science Foundation of China (No. 62372271 and No. 62595772) and the projects of Beijing Science and Technology Program (Z251100008125028).

%% file: sec/X_suppl.tex
\clearpage
\begin{center}
    {\Large\bfseries Supplementary Material}
\end{center}
\vspace{1em}

\appendix
\setcounter{table}{0}
\setcounter{figure}{0}
\setcounter{equation}{0}
\renewcommand{\thetable}{A\arabic{table}}
\renewcommand{\thefigure}{A\arabic{figure}}
\renewcommand{\theequation}{A\arabic{equation}}
\renewcommand{\theHsection}{supp.\arabic{section}}
\renewcommand{\theHsubsection}{supp.\arabic{section}.\arabic{subsection}}
\renewcommand{\theHtable}{supp.\arabic{table}}
\renewcommand{\theHfigure}{supp.\arabic{figure}}
\renewcommand{\theHequation}{supp.\arabic{equation}}

\section{Appendix Overview}
In this appendix, we provide additional details that were not covered in the main text. In \sec~\ref{sec:supp:implementation}, we describe several implementation aspects of RefracGS, including the architecture of the neural height-field network and the derivation of gradient formulas for computing the loss with respect to the ray origin and direction. In \sec~\ref{sec:supp:dataset}, we present a comprehensive description of the RefracGS dataset, covering scene and camera configurations as well as a comparison with the NeRFrac dataset. In \sec~\ref{sec:supp:experiments}, we supplement the experimental section with further information, including the runtime breakdown of our method, the parameter settings used for NeRFrac and evaluations of refractive object reconstruction methods. In \sec~\ref{sec:supp:limitations}, we present a thorough discussion regarding the current limitations of our method along with potential directions for future work. In \sec~\ref{sec:supp:results}, we report additional experimental results.

\section{Implementation Details}
\label{sec:supp:implementation}
In this section, we supplement the implementation details that were not covered in the main paper. We first provide a detailed description of the architecture of our neural water surface representation, followed by a complete derivation of the backward gradients used to optimize refractive surfaces.
\subsection{Implementation of Water Height Map}
Our height-field MLP consists of six hidden layers, each with 256 neurons and Leaky ReLU activations. We apply the same frequency encoding to the input coordinates as used in NeRF~\cite{mildenhall2020nerf}, with the number of frequency bands empirically set to 6. For the ray-water surface intersection, we initialize with a $200 \times 200$ coarse grid for proxy mesh and empirically execute 2 iterations of recursive subdivision. The entire network is implemented in PyTorch.

\subsection{Gradient for Ray Origin and Direction}
To enable the color loss to effectively guide the optimization of the refractive surface geometry, we need to compute the gradients of the loss function $L$ with respect to the origin of the refracted ray origin $\textbf{p}$ (i.e., the ray–surface intersection) and refracted direction $\mathbf{\omega_o}$. For $\textbf{p}$, we have:
\begin{equation}
\label{eq:gradient:position}
\frac{\partial L}{\partial \mathbf{p}}=\frac{\partial L}{\partial \mathbf{C}} \cdot \frac{\partial \mathbf{C}}{\partial \mathbf{p}}=\frac{\partial L}{\partial \mathbf{C}} \cdot \sum_{k=1}^N (\frac{\partial \textbf{C}}{\partial \alpha_k}\cdot \frac{\partial \alpha_k}{\partial \textbf{p}} + \frac{\partial \textbf{C}}{\partial \textbf{c}_k}\cdot \frac{\partial \textbf{c}_k}{\partial \textbf{p}})
\end{equation}
where $\textbf{C}$ denotes the final color accumulated along the current ray through alpha blending, while $\alpha_k$ and $\textbf{c}_k$ represent the density and color, respectively, of the $k$-th intersected Gaussian at the corresponding intersection point. Similarly, for $\mathbf{\omega_o}$, we have:

\begin{equation}
\label{eq:gradient:direction}
\frac{\partial L}{\partial \mathbf{\omega_o}}=\frac{\partial L}{\partial \mathbf{C}} \cdot \sum_{k=1}^N (\frac{\partial \textbf{C}}{\partial \alpha_k}\cdot \frac{\partial \alpha_k}{\partial \mathbf{\omega_o}} + \frac{\partial \textbf{C}}{\partial \textbf{c}_k}\cdot \frac{\partial \textbf{c}_k}{\partial \mathbf{\omega_o}})
\end{equation}

Similar to the computation in 3DGRT~\cite{loccoz20243dgrt}, for each intersection between the ray and each Gaussian, the alpha blending formula can be decomposed into the following form:
\begin{equation}
\mathbf{C}=\sum_{i=1}^{k-1}\alpha_i\textbf{c}_iT_i + \alpha_k\textbf{c}_kT_k + (1-\alpha_k)T_k\sum_{i=k+1}^N \alpha_i\textbf{c}_iT_i
\end{equation}
thus, we can obtain the two components $\frac{\partial \textbf{C}}{\partial \alpha_k}$ and $\frac{\partial \textbf{C}}{\partial \textbf{c}_k}$ in \eq~\ref{eq:gradient:position} and \eq~\ref{eq:gradient:direction} as follows:

\begin{equation}
\frac{\partial \textbf{C}}{\partial \alpha_k} = \textbf{c}_kT_k - T_k\sum_{i=k+1}^N \alpha_i\textbf{c}_iT_i
\end{equation}

\begin{equation}
\frac{\partial \textbf{C}}{\partial \textbf{c}_k} = \alpha_kT_k
\end{equation}

At each ray-Gaussian intersection, the color is computed via spherical harmonics $\textbf{c}_k=\text{SH}(\mathbf{\omega_o})$. We use $\text{SH}_{bwd}$ to represent the backward propagation function of the spherical harmonics function and thus we have:
\begin{equation}
\frac{\partial \textbf{c}_k}{\partial \mathbf{\omega_o}} = \text{SH}_{bwd}(\mathbf{\omega_o})
\end{equation}
\begin{equation}
\frac{\partial \textbf{c}_k}{\partial \textbf{p}}=0
\end{equation}

Following 3DGRT, we use the maximum density-response point of each Gaussian on the ray to determine the intersection position. So we have:
\begin{equation}
\alpha_k = o_ke^{-\frac{\|\mathbf{p}_g \times \mathbf{\omega_g}\|^2}{2}}
\end{equation}
where $o_k$ denotes the density attribute of the $k$-th Gaussian, $\mathbf{p}_g=\mathbf{S}_k^{-1}\mathbf{R}_k^T(\mathbf{p}-\mathbf{\mu}_k)$ and $\mathbf{\omega_g}=\mathbf{S}_k^{-1}\mathbf{R}_k^T\mathbf{\omega_o}$. Thus, we can obtain the rest two components $\frac{\partial \alpha_k}{\partial \mathbf{p}}$ and $\frac{\partial \alpha_k}{\partial \mathbf{\omega_o}}$ in \eq~\ref{eq:gradient:position} and \eq~\ref{eq:gradient:direction} as follows:

\begin{equation}
\frac{\partial \alpha_k}{\partial \mathbf{p}} = \frac{\partial \alpha_k}{\partial \mathbf{p}_g} \cdot  \frac{\partial \mathbf{p}_g}{\partial \mathbf{p}} = -\alpha_k\Big( \|\mathbf{\omega_g}\|^2\,\mathbf{p}_g - (\mathbf{p}_g\cdot\mathbf{\omega_g})\,\mathbf{\omega_g} \Big) \cdot \frac{\partial \mathbf{p}_g}{\partial \mathbf{p}}
\end{equation}
\begin{equation}
\frac{\partial \alpha_k}{\partial \mathbf{\omega_o}} = \frac{\partial \alpha_k}{\partial \mathbf{\omega_g}} \cdot \frac{\partial \mathbf{\omega_g}}{\partial \mathbf{\omega_o}} = -\alpha_k\Big( \|\mathbf{p}_g\|^2\,\mathbf{\omega_g} - (\mathbf{p}_g\cdot\mathbf{\omega_g})\,\mathbf{p}_g \Big) \cdot \frac{\partial \mathbf{\omega_g}}{\partial \mathbf{\omega_o}}
\end{equation}
\begin{equation}
\frac{\partial \mathbf{p}_g}{\partial \mathbf{p}} = \frac{\partial \mathbf{\omega_g}}{\partial \mathbf{\omega_o}} =\mathbf{S}_k^{-1}\mathbf{R}_k^T
\end{equation}

The above derivation allows the gradient of the loss function to propagate to both the ray origin and direction, enabling the optimization of scene parameters that affect ray trajectories. Our experiments further demonstrate the feasibility of reconstructing refractive surfaces using this approach.

\section{Dataset Details}
\label{sec:supp:dataset}
In this section, we provide a detailed description of the RefracGS dataset and present a comparative analysis with the NeRFrac dataset.
\subsection{Scenes Setting}
The RefracGS dataset comprises three carefully designed scenes, each spanning an approximate volume of $1.5\text{m} \times 1.5\text{m} \times 1\text{m}$ and built from openly licensed model assets. We employed fluid simulation to construct water surfaces with irregular ripples, with all objects placed below the surface and all cameras positioned above it. Since our study focuses exclusively on the refractive effects through surfaces, we eliminated reflection effects and retained only refraction.

\subsection{Cameras Setting}
In the RefracGS dataset, all cameras are distributed on a hemispherical surface centered at the scene origin and oriented downward. The training set consists of 24 views, evenly arranged along two concentric rings with elevation angles of 30° and 45° relative to the z-axis. The testing set comprises 6 views, distributed along a single ring at a 35° elevation, also directed downward toward the scene.

\begin{figure}[htbp]
    \centering
    \begin{minipage}{0.45\textwidth}
        \centering
        \includegraphics[height=0.55\linewidth]{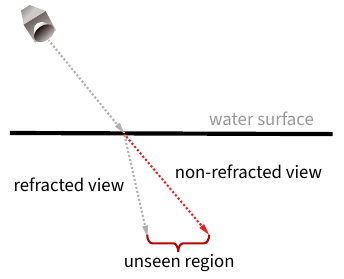}
        \caption{\textbf{The refractive effect of the water surface reduces the visible range compared to air.} 
        Directly rendering the test views in air would cause them to include regions not covered by the training views, thereby making the results unreliable.}
        \label{fig:supp:unseen}
    \end{minipage}
    \hspace{0.05\textwidth}
    \begin{minipage}{0.45\textwidth}
        \centering
        \begin{subfigure}[t]{0.48\textwidth}
            \centering
            \includegraphics[height=\linewidth]{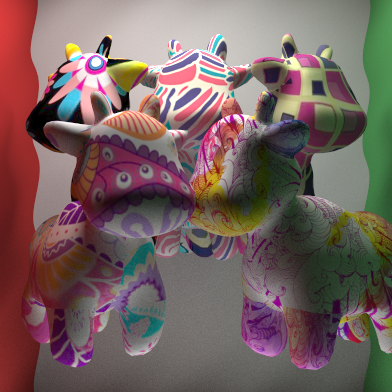}
            \caption{Ground Truth}
            \label{fig:supp:nunerf_a}
        \end{subfigure}
        \hfill
        \begin{subfigure}[t]{0.48\textwidth}
            \centering
            \includegraphics[height=\linewidth]{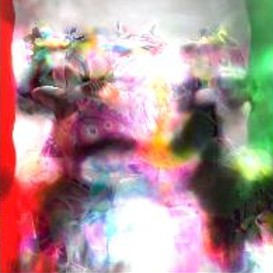}
            \caption{NU-NeRF}
            \label{fig:supp:nunerf_b}
        \end{subfigure}
        \caption{Comparison between \textbf{(a) the ground truth} and \textbf{(b) the NU-NeRF Stage 1 training results} on the NeRFrac synthetic dataset.
        In refractive water-surface scenes, NU-NeRF fails to correctly reconstruct the refractive interface during training.}
        \label{fig:supp:nunerf}
    \end{minipage}
    \vspace{-4mm}
\end{figure}

Directly using the camera poses of the test views to evaluate water removal exposes large regions of the scene that are not observed during training, thereby making the results unreliable, as illustrated in \fig~\ref{fig:supp:unseen}. 
To address this issue, we additionally provide six water removal test views to ensure that all tested regions are fully covered by the training views. These views are arranged in a circular configuration at a 30° elevation relative to the z-axis, but positioned closer to the scene center in order to mitigate the adverse effects of visibility changes caused by refraction.

\subsection{Comparing with NeRFrac Dataset}
\label{sec:supp:camera9}
In the NeRFrac dataset, each scene provides nine views, where a single test view is surrounded by eight training views, all oriented vertically downward. In contrast, the RefracGS dataset adopts more oblique camera angles and increases the number of views to ensure sufficient coverage. This oblique arrangement allows the capture of richer side information of objects, thereby enabling the reconstruction of underwater geometry. By comparison, the NeRFrac dataset lacks adequate side-view information and thus fails to recover complete underwater object geometry. Our experiments demonstrate that the proposed camera configuration is sufficient to reconstruct the full geometry of underwater objects, as illustrated in \fig 1 of the main text.

However, increasing the camera tilt angle enlarges the incidence angle of light, thereby amplifying the refractive effects. As a result, the RefracGS dataset is more challenging than the NeRFrac dataset. Our experimental results show that all existing methods, including NeRFrac, fail to produce reliable reconstructions under such conditions, whereas our approach is able to achieve high-quality reconstructions.

In addition, we provide a camera configuration in the RefracGS scenes similar to that of the NeRFrac dataset, consisting of nine cameras, all oriented vertically downward toward the scene. As shown in \tab~\ref{tab:supp:refracgs9}, our experiments demonstrate that NeRFrac achieves relatively good performance under this setting; however, our method still delivers a substantial improvement in reconstruction quality.

\vspace{-2mm}
\section{Experiments Details}
\label{sec:supp:experiments}
\subsection{Runtime Breakdown}
To evaluate the computational efficiency of our method, we provide a detailed runtime breakdown for both the training and inference phases. 
During training, the overall computation comprises coarse proxy mesh and BVH construction (13\%), global intersection testing (2\%), recursive subdivision tracing (74\%), and Gaussian rendering (11\%). 
For inference, because the optimized surface is fixed, our method eliminates the need for repeated proxy mesh and BVH rebuilding. Consequently, the inference pipeline requires only a single high-resolution global intersection test (31\%) and Gaussian rendering (69\%).

\subsection{Parameter Setting for NeRFrac}
In the released NeRFrac code, two parameter settings are provided: for the real dataset, zero-order frequency encoding with 200,000 training iterations is used, whereas for the synthetic dataset, eighth-order frequency encoding with 100,000 iterations is adopted. This accounts for the nearly twofold difference in training time between the two datasets. For the RefracGS dataset, we consistently observe that eighth-order frequency encoding outperforms zero-order encoding, and thus we uniformly adopt the eighth-order setting.  Due to the larger number of input views, we employ 200,000 iterations. Nevertheless, we find that NeRFrac failed to obtain reasonable results under both training schedules. Under the additional 9-view configuration in RefracGS scenes, we follow the synthetic setup with 100,000 training iterations, as further increasing the iteration count yields no improvement.

\subsection{Refractive Object Reconstruction Method}
As noted in the Related Work section, several methods focus on reconstructing solid refractive objects in air. To examine whether these approaches can be extended to scene through refractive surfaces, we evaluated NU-NeRF~\cite{Jia2024NUNeRF} on the NeRFrac~\cite{zhan2023nerfrac} dataset. We found that, under through-surface refraction settings, NU-NeRF consistently fails to reconstruct the refractive surfaces, as illustrated in \fig~\ref{fig:supp:nunerf}. This observation suggests that these methods are not well-suited to our scene configuration.

\begin{wrapfigure}{r}{0.5\textwidth}
    \centering
    \vspace{-8mm}
 
    \begin{subfigure}[b]{0.16\textwidth}
        \centering
        \includegraphics[width=\textwidth]{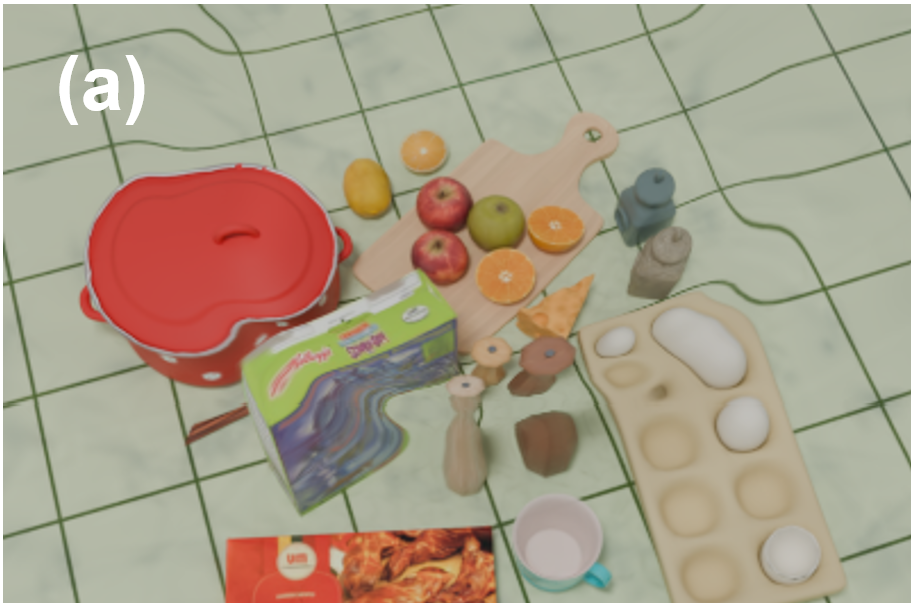}
        \label{fig:sub1}
    \end{subfigure}
    \hfill 
    \begin{subfigure}[b]{0.16\textwidth}
        \centering
        \includegraphics[width=\textwidth]{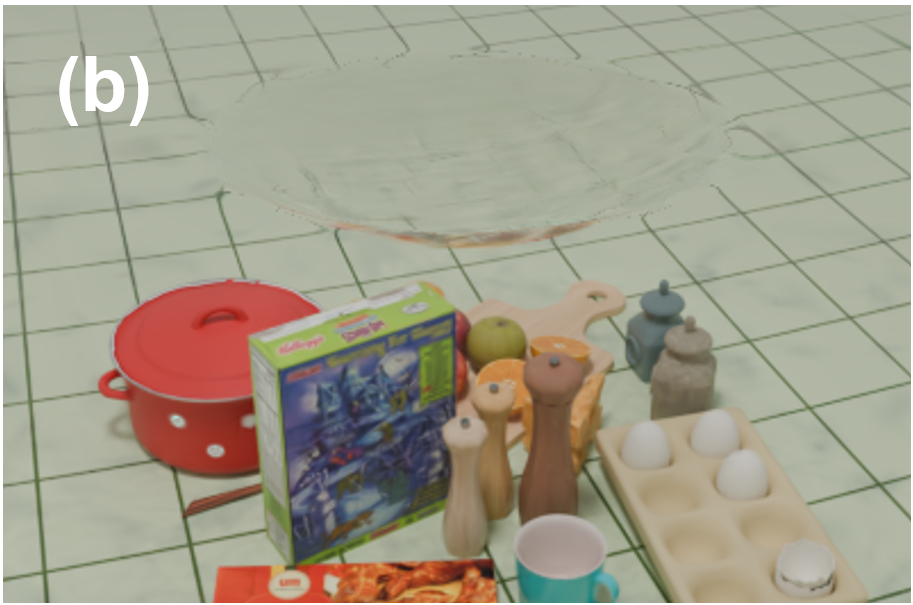}
        \label{fig:sub2}
    \end{subfigure}
    \hfill
    \begin{subfigure}[b]{0.16\textwidth}
        \centering
        \includegraphics[width=\textwidth]{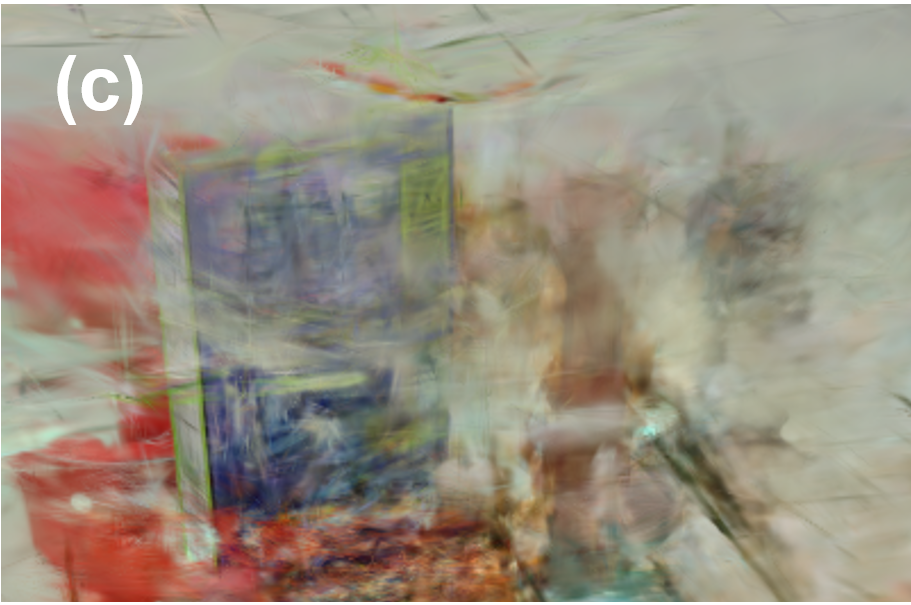}
        \label{fig:sub3}
    \end{subfigure}
    
    \vspace{-6mm}
    \caption{\textbf{Boundary cases of single-refraction.} (a) When cameras are fully above the water, the rays are downward or slightly oblique, single refraction holds. (b) When cameras are near water, extreme grazing angles cause multiple refractions. (c) Submerged cameras are more complicated and are beyond models of our method.}
    \label{fig:supp:limitation}
    \vspace{-12mm}
\end{wrapfigure}

\section{Discussion about Limitations}
\label{sec:supp:limitations}

In this section, we provide a comprehensive discussion regarding the current limitations of our method and outline several promising directions for future exploration.

\paragraph{Radiometric Effects.}
Our current framework primarily focuses on resolving geometric distortions induced by refraction under the water surface, which remains a challenging and under-explored problem in multi-view reconstruction. Consequently, complex radiometric effects such as surface reflections and volumetric scattering are omitted in our current implementation. Nevertheless, these phenomena are orthogonal to our geometric formulation and can be incrementally integrated into our pipeline. For instance, surface reflection could be effectively modeled via Fresnel-weighted ray mixing, whereas volumetric scattering could be physically accumulated during the alpha blending phase along the ray casting paths.

\paragraph{Dynamic Water Surfaces.}
Our method is designed for scenarios where the undulating water surface is captured within a single-frame. Extending the current formulation to dynamic, multi-frame aquatic environments represents a natural and crucial next step. Our framework can be gracefully extended to time-varying water surfaces by incorporating a temporal variable ($t$) as an additional input to the water surface MLP. This would allow the network to learn a continuous, time-dependent deformation of the fluid interface across consecutive frames.

\paragraph{Single-Value Height Field Assumption.}
Our water surface representation relies on a single-value height field assumption, which is specifically optimized for downward-facing viewpoints. While this assumption holds true and remains sufficient for the vast majority of scenarios, it introduces inherent limitations in extreme geometric topologies. Specifically, it fails under highly oblique viewing angles or complex fluid dynamics such as curling crests and breaking waves. We illustrate these failure cases and extreme configurations on our synthetic dataset in Fig.~\ref{fig:supp:limitation}, demonstrating the bounds of the height field assumption.

\paragraph{Underwater Representation Choice.}
To achieve high-fidelity rendering and novel view synthesis, we adopt 3D Gaussian Splatting (3DGS) as our standard representation for the underwater scene. However, while 3DGS excels at preserving visual appearance, it lacks explicit surface constraints, which may compromise exact geometric reconstruction. For applications that require precise 3D mesh extraction or strict structural geometry, our refractive ray tracing formulation remains fully compatible with alternative scene representations, such as 2D Gaussian Splatting (2DGS)~\cite{Huang2DGS2024} or neural implicit surfaces~\cite{wang2021neus}, without requiring any modifications to the water surface modeling.

\section{More Results}
\label{sec:supp:results}
In this section, we provide additional experimental results on the Nerfrac real (\tab~\ref{tab:supp:nerfracreal}, \fig~\ref{fig:supp:nerfracreal}) and synthetic (\tab~\ref{tab:supp:nerfracsyn}, \fig~\ref{fig:supp:nerfracsyn}) dataset, as well as the RefracGS dataset (\tab~\ref{tab:supp:refracgs}, \fig~\ref{fig:supp:refracgs}). Moreover, as discussed in \sec~\ref{sec:supp:camera9}, we further evaluate all methods under the 9-view setting on the RefracGS scenes (\tab~\ref{tab:supp:refracgs9}, \fig~\ref{fig:supp:refracgs9}). We compare the performance of Mip-NeRF~\cite{barron2021mipnerf}, TensoRF~\cite{Chen2022tensorf}, Plenoxels~\cite{yu2022plenoxels}, 3DGS~\cite{kerbl3Dgaussians}, 3DGRT~\cite{loccoz20243dgrt}, WSGS~\cite{yoon2026wsgs}, NeRFrac~\cite{zhan2023nerfrac} and our proposed method. Additionally, we further evaluate 3DGRT and 3DGS under a training schedule of 15,000 iterations to align with our method.

\clearpage
\begin{table}[htbp]
  \centering
  \caption{Quantitative results on the Nerfrac real dataset. (* indicates models trained for 15,000 iterations; $\sim$ indicates that the runtime was not precisely recorded for each scene, but rather estimated. $\dagger$ indicates methods specifically designed for refractive scenes.)
}
  \vspace{2pt}
  \label{tab:supp:nerfracreal}

  \setlength{\tabcolsep}{8pt}

  \resizebox{0.93\linewidth}{!}{%
  \begin{tabular}{ll ccccc}
    \toprule
    \textbf{Scene} & \textbf{Method} & \textbf{PSNR\textuparrow{}} & \textbf{SSIM\textuparrow{}} & \textbf{LPIPS\textdownarrow{}} &
    \textbf{Train\textdownarrow{}} &
    \textbf{FPS\textuparrow{}}\\
    \midrule
	\multirow{9}{*}{Plant}
	& Mip-NeRF & 10.350 & 0.322 & 0.945 & $\sim$15 h & $<$1 \\
    & TensoRF & 21.857 & 0.798 & 0.138 & $\sim$11 min & $<$1 \\
    & Plenoxels & 14.491 & 0.378 & 0.655 & $\sim$9 min & $\sim$60 \\
    & 3DGS & 28.520 & 0.801 & 0.198 & 3 min & \textbf{1019} \\
    & 3DGRT & 27.579 & 0.845 & 0.155 & 13 min & 143 \\
    & 3DGS* & 27.007 & 0.870 & 0.136 & \textbf{2 min} & 932 \\
    & 3DGRT* & 27.661 & 0.856 & 0.146 & 7 min & 124 \\
    & WSGS$\dagger$ & 18.742 & 0.594 & 0.375 & 106 min & 131 \\
    & NeRFrac$\dagger$ & 28.413 & 0.879 & 0.156 & 150 min & $<$1 \\
    & Ours$\dagger$ & \textbf{30.729} & \textbf{0.922} & \textbf{0.120} & 10 min & 223 \\
 
    \midrule
	\multirow{9}{*}{Tree}
	& Mip-NeRF & 17.386 & 0.464 & 0.875 & $\sim$15 h & $<$1 \\
    & TensoRF & 25.235 & 0.772 & 0.190 & $\sim$11 min & $<$1 \\
    & Plenoxels & 18.476 & 0.417 & 0.628 & $\sim$9 min & $\sim$60 \\
    & 3DGS & 26.751 & 0.859 & \textbf{0.144} & 3 min & 843 \\
    & 3DGRT & 28.292 & 0.776 & 0.260 & 16 min & 259 \\
    & 3DGS* & 28.670 & 0.806 & 0.193 & \textbf{2 min} & \textbf{1034} \\
    & 3DGRT* & 28.369 & 0.782 & 0.246 & 7 min & 222 \\
    & WSGS$\dagger$ & 22.400 & 0.463 & 0.443 & 105 min & 127 \\
    & NeRFrac$\dagger$ & 30.456 & 0.857 & 0.221 & 149 min & $<$1 \\
    & Ours$\dagger$ & \textbf{30.565} & \textbf{0.865} & 0.169 & 10 min & 291 \\

    \midrule
	\multirow{9}{*}{Fish}
	& Mip-NeRF & 8.701 & 0.257 & 0.976 & $\sim$15 h & $<$1 \\
    & TensoRF & 22.094 & 0.858 & 0.097 & $\sim$11 min & $<$1 \\
    & Plenoxels & 11.064 & 0.361 & 0.678 & $\sim$9 min & $\sim$60 \\
    & 3DGS & 28.101 & 0.897 & 0.096 & 3 min & 761 \\
    & 3DGRT & 27.809 & 0.868 & 0.128 & 18 min & 120 \\
    & 3DGS* & 28.188 & 0.906 & \textbf{0.088} & \textbf{2 min} & \textbf{887} \\
    & 3DGRT* & 27.878 & 0.882 & 0.124 & 7 min & 80 \\
    & WSGS$\dagger$ & 26.276 & 0.891 & 0.104 & 111 min & 129 \\
    & NeRFrac$\dagger$ & 27.183 & 0.891 & 0.135 & 146 min & $<$1 \\
    & Ours$\dagger$ & \textbf{31.859} & \textbf{0.945} & 0.117 & 10 min & 194 \\
    \midrule
    
	\multirow{9}{*}{Redflower}
	& Mip-NeRF & 10.385 & 0.221 & 0.916 & $\sim$15 h & $<$1 \\
    & TensoRF & 22.709 & 0.809 & 0.133 & $\sim$11 min & $<$1 \\
    & Plenoxels & 15.437 & 0.403 & 0.566 & $\sim$9 min & $\sim$60 \\
    & 3DGS & 27.475 & 0.900 & 0.082 & 3 min & 844 \\
    & 3DGRT & 27.251 & 0.881 & 0.113 & 19 min & 158 \\
    & 3DGS* & 27.741 & 0.906 & \textbf{0.078} & \textbf{2 min} & \textbf{854} \\
    & 3DGRT* & 27.192 & 0.887 & 0.106 & 7 min & 133 \\
    & WSGS$\dagger$ & 22.376 & 0.767 & 0.193 & 103 min & 129 \\
    & NeRFrac$\dagger$ & 26.535 & 0.880 & 0.157 & 153 min & $<$1 \\
    & Ours$\dagger$ & \textbf{29.533} & \textbf{0.921} & 0.110 & 10 min & 287 \\
    \bottomrule
  \end{tabular}%
  }
\end{table}

\begin{table}[htbp]
  \centering
  \caption{Quantitative results on the Nerfrac synthetic dataset. (* indicates models trained for 15,000 iterations; $\sim$ indicates that the runtime was not precisely recorded for each scene, but rather estimated. $\dagger$ indicates methods specifically designed for refractive scenes.)
}
  \vspace{2pt}
  \label{tab:supp:nerfracsyn}

  \setlength{\tabcolsep}{8pt}

  \resizebox{0.93\linewidth}{!}{%
  \begin{tabular}{ll ccccc}
    \toprule
    \textbf{Scene} & \textbf{Method} & \textbf{PSNR\textuparrow{}} & \textbf{SSIM\textuparrow{}} & \textbf{LPIPS\textdownarrow{}} &
    \textbf{Train\textdownarrow{}} &
    \textbf{FPS\textuparrow{}}\\
    \midrule
	\multirow{9}{*}{Primary Sine}
	& Mip-NeRF & 12.482 & 0.402 & 0.852 & $\sim$15 h & $<$1 \\
    & TensoRF & 16.301 & 0.527 & 0.614 & $\sim$11 min & $<$1 \\
    & Plenoxels & 13.390 & 0.403 & 0.810 & $\sim$12 min & $\sim$70 \\
    & 3DGS & 21.279 & 0.705 & 0.311 & 3 min & \textbf{693} \\
    & 3DGRT & 21.135 & 0.663 & 0.390 & 29 min & 80 \\
    & 3DGS* & 21.459 & 0.726 & 0.346 & \textbf{2 min} & 686 \\
    & 3DGRT* & 21.514 & 0.690 & 0.378 & 9 min & 89 \\
    & WSGS$\dagger$ & 20.032 & 0.679 & 0.318 & 109 min & 157 \\
    & NeRFrac$\dagger$ & 34.332 & 0.944 & 0.150 & 78 min & $<$1 \\
    & Ours$\dagger$ & \textbf{35.391} & \textbf{0.951} & \textbf{0.136} & 14 min & 216 \\
     
    \midrule
	\multirow{9}{*}{Second Sine}
	& Mip-NeRF & 13.088 & 0.373 & 0.877 & $\sim$15 h & $<$1 \\
    & TensoRF & 16.246 & 0.545 & 0.568 & $\sim$11 min & $<$1 \\
    & Plenoxels & 12.901 & 0.433 & 0.811 & $\sim$12 min & $\sim$70 \\
    & 3DGS & 24.962 & 0.825 & 0.221 & 3 min & \textbf{747} \\
    & 3DGRT & 24.089 & 0.773 & 0.287 & 22 min & 108 \\
    & 3DGS* & 23.954 & 0.816 & 0.251 & \textbf{2 min} & 713 \\
    & 3DGRT* & 23.942 & 0.793 & 0.280 & 8 min & 64 \\
    & WSGS$\dagger$ & 22.300 & 0.804 & 0.197 & 111 min & 152 \\
    & NeRFrac$\dagger$ & 35.307 & 0.949 & 0.141 & 78 min & $<$1 \\
    & Ours$\dagger$ & \textbf{36.964} & \textbf{0.956} & \textbf{0.133} & 14 min & 197 \\

    \midrule
	\multirow{9}{*}{Hybrid}
	& Mip-NeRF & 11.672 & 0.371 & 0.881 & $\sim$15 h & $<$1 \\
    & TensoRF & 16.123 & 0.525 & 0.622 & $\sim$11 min & $<$1 \\
    & Plenoxels & 13.711 & 0.386 & 0.798 & $\sim$12 min & $\sim$70 \\
    & 3DGS & 20.319 & 0.662 & 0.370 & 4 min & 673 \\
    & 3DGRT & 20.564 & 0.633 & 0.413 & 30 min & 84 \\
    & 3DGS* & 20.679 & 0.684 & 0.374 & \textbf{2 min} & \textbf{724} \\
    & 3DGRT* & 20.629 & 0.658 & 0.413 & 9 min & 88 \\
    & WSGS$\dagger$ & 20.273 & 0.694 & 0.299 & 103 min & 149 \\
    & NeRFrac$\dagger$ & 33.504 & 0.939 & 0.158 & 77 min & $<$1 \\
    & Ours$\dagger$ & \textbf{35.093} & \textbf{0.949} & \textbf{0.139} & 14 min & 196 \\
    \bottomrule
  \end{tabular}%
  }
\end{table}

\begin{table}[htbp]
  \centering
  \caption{Quantitative results on the RefracGS dataset. (* indicates models trained for 15,000 iterations; $\sim$ indicates that the runtime was not precisely recorded for each scene, but rather estimated. $\dagger$ indicates methods specifically designed for refractive scenes.)
}
  \vspace{2pt}
  \label{tab:supp:refracgs}

  \setlength{\tabcolsep}{5pt}

  \resizebox{0.93\linewidth}{!}{%
  \begin{tabular}{ll cccccc}
    \toprule
    \textbf{Scene} & \textbf{Method} & \textbf{PSNR\textuparrow{}} & \textbf{SSIM\textuparrow{}} & \textbf{LPIPS\textdownarrow{}} &
    \textbf{Train\textdownarrow{}} &
    \textbf{FPS\textuparrow{}} &
    \textbf{RMSE(cm)\textdownarrow{}}\\
    \midrule
	\multirow{9}{*}{Toys}
	& Mip-NeRF & 12.122 & 0.363 & 0.855 & $\sim$15 h & $<$1 & - \\
    & TensoRF & 12.353 & 0.255 & 0.621 & $\sim$13 min & $<$1 & - \\
    & Plenoxels & 12.616 & 0.286 & 0.717 & $\sim$12 min & $\sim$20 & - \\
    & 3DGS & 15.230 & 0.297 & 0.525 & 5 min & \textbf{526} & - \\
    & 3DGRT & 16.027 & 0.304 & 0.566 & 16 min & 171 & - \\
    & 3DGS* & 15.085 & 0.303 & 0.528 & \textbf{2 min} & 425 & - \\
    & 3DGRT* & 16.175 & 0.328 & 0.548 & 7 min & 141 & - \\
    & WSGS$\dagger$ & 14.599 & 0.285 & 0.532 & 106 min & 127 \\
    & NeRFrac$\dagger$ & 16.902 & 0.321 & 0.644 & 165 min & $<$1 & 3.850 \\
    & Ours$\dagger$ & \textbf{29.638} & \textbf{0.937} & \textbf{0.068} & 11 min & 122 & \textbf{0.055} \\

    \midrule
	\multirow{9}{*}{Desktop}
	& Mip-NeRF & 14.091 & 0.526 & 0.743 & $\sim$15 h & $<$1 & - \\
    & TensoRF & 14.432 & 0.434 & 0.656 & $\sim$13 min & $<1$ & - \\
    & Plenoxels & 13.798 & 0.438 & 0.677 & $\sim$12 min & $\sim$20 & - \\
    & 3DGS & 17.952 & 0.511 & 0.459 & 5 min & 596 & - \\
    & 3DGRT & 18.078 & 0.490 & 0.535 & 14 min & 160 & - \\
    & 3DGS* & 18.179 & 0.525 & 0.449 & \textbf{2 min} & \textbf{640} & - \\
    & 3DGRT* & 18.413 & 0.521 & 0.514 & 7 min & 100 & - \\
    & WSGS$\dagger$ & 20.625 & 0.474 & 0.459 & 107 min & 123 \\
    & NeRFrac$\dagger$ & 17.102 & 0.443 & 0.611 & 159 min & $<$1 & 3.848 \\
    & Ours$\dagger$ & \textbf{30.792} & \textbf{0.926} & \textbf{0.102} & 11 min & 119 & \textbf{0.177} \\

    \midrule
	\multirow{9}{*}{Kitchen}
	& Mip-NeRF & 15.411 & 0.550 & 0.827 & $\sim$15 h & $<$1 & - \\
    & TensoRF & 15.592 & 0.399 & 0.699 & $\sim$13 min & $<$1 & - \\
    & Plenoxels & 15.535 & 0.425 & 0.734 & $\sim$12 min & $\sim$20 & - \\
    & 3DGS & 19.265 & 0.477 & 0.475 & 5 min & \textbf{673} & - \\
    & 3DGRT & 19.953 & 0.480 & 0.534 & 12 min & 227 & - \\
    & 3DGS* & 19.003 & 0.472 & 0.493 & \textbf{2 min} & 601 & - \\
    & 3DGRT* & 19.839 & 0.489 & 0.529 & 6 min & 161 & - \\
    & WSGS$\dagger$ & 18.523 & 0.513 & 0.463 & 103 min & 122 \\
    & NeRFrac$\dagger$ & 19.508 & 0.451 & 0.618 & 168 min & $<$1 & 3.267 \\
    & Ours$\dagger$ & \textbf{30.244} & \textbf{0.936} & \textbf{0.124} & 11 min & 131 & \textbf{0.115} \\
     
    \bottomrule
  \end{tabular}%
  }
\end{table}

\begin{table}[htbp]
  \centering
  \caption{Quantitative results of different methods on RefracGS dataset scenes under the additional nine-view setting as discussed in \sec~\ref{sec:supp:camera9}. (* indicates models trained for 15,000 iterations; $\sim$ indicates that the runtime was not precisely recorded for each scene, but rather estimated. $\dagger$ indicates methods specifically designed for refractive scenes.)
}
  \vspace{2pt}
  \label{tab:supp:refracgs9}

  \setlength{\tabcolsep}{5pt}

  \resizebox{0.93\linewidth}{!}{%
  \begin{tabular}{ll cccccc}
    \toprule
    \textbf{Scene} & \textbf{Method} & \textbf{PSNR\textuparrow{}} & \textbf{SSIM\textuparrow{}} & \textbf{LPIPS\textdownarrow{}} &
    \textbf{Train\textdownarrow{}} &
    \textbf{FPS\textuparrow{}} &
    \textbf{RMSE(cm)\textdownarrow{}}\\
    \midrule
	\multirow{9}{*}{Toys}
	& Mip-NeRF & 12.637 & 0.290 & 0.833 & $\sim$15 h & $<$1 & - \\
    & TensoRF & 13.657 & 0.287 & 0.519 & $\sim$13 min & $<$1 & - \\
    & Plenoxels & 12.261 & 0.271 & 0.701 & $\sim$12 min & $\sim$20 & - \\
    & 3DGS & 18.840 & 0.536 & 0.352 & 3 min & \textbf{625} & - \\
    & 3DGRT & 19.646 & 0.562 & 0.394 & 23 min & 114 & - \\
    & 3DGS* & 19.398 & 0.559 & 0.333 & \textbf{2 min} & 621 & - \\
    & 3DGRT* & 19.673 & 0.570 & 0.387 & 9 min & 108 & - \\
    & WSGS$\dagger$ & 19.671 & 0.559 & 0.359 & 166 min & 108 & - \\
    & NeRFrac$\dagger$ & 25.673 & 0.846 & 0.186 & 75 min & $<$1 & 8.664 \\
    & Ours$\dagger$ & \textbf{28.155} & \textbf{0.904} & \textbf{0.113} & 15 min & 184 & \textbf{1.394} \\

    \midrule
	\multirow{9}{*}{Desktop}
	& Mip-NeRF & 9.425 & 0.267 & 0.862 & $\sim$15 h & $<$1 & - \\
    & TensoRF & 16.297 & 0.465 & 0.617 & $\sim$13 min & $<$1 & - \\
    & Plenoxels & 14.069 & 0.385 & 0.670 & $\sim$12 min & $\sim$20 & - \\
    & 3DGS & 21.846 & 0.681 & 0.276 & 3 min & 752 & - \\
    & 3DGRT & 22.538 & 0.698 & 0.323 & 19 min & 133 & - \\
    & 3DGS* & 22.054 & 0.694 & 0.267 & \textbf{2 min} & \textbf{871} & - \\
    & 3DGRT* & 22.663 & 0.698 & 0.318 & 9 min & 91 & - \\
    & WSGS$\dagger$ & 21.996 & 0.695 & 0.295 & 162 min & 94 & - \\
    & NeRFrac$\dagger$ & 29.063 & 0.886 & 0.172 & 75 min & $<$1 & 5.341 \\
    & Ours$\dagger$ & \textbf{34.034} & \textbf{0.961} & \textbf{0.079} & 15 min & 155 & \textbf{1.508} \\

    \midrule
	\multirow{9}{*}{Kitchen}
	& Mip-NeRF & 15.387 & 0.459 & 0.919 & $\sim$15 h & $<$1 & - \\
    & TensoRF & 18.510 & 0.505 & 0.558 & $\sim$13 min & $<$1 & - \\
    & Plenoxels & 15.987 & 0.480 & 0.729 & $\sim$12 min & $\sim$20 & - \\
    & 3DGS & 22.540 & 0.679 & 0.309 & 3 min & 758 & - \\
    & 3DGRT & 23.173 & 0.695 & 0.365 & 16 min & 116 & - \\
    & 3DGS* & 22.689 & 0.692 & 0.303 & \textbf{2 min} & \textbf{771} & - \\
    & 3DGRT* & 23.078 & 0.700 & 0.351 & 8 min & 124 & - \\
    & WSGS$\dagger$ & 23.240 & 0.699 & 0.317 & 172 min & 87 & - \\
    & NeRFrac$\dagger$ & 26.222 & 0.854 & 0.233 & 75 min & $<$1 & 6.367 \\
    & Ours$\dagger$ & \textbf{29.939} & \textbf{0.933} & \textbf{0.144} & 15 min & 124 & \textbf{1.996} \\
 
    \bottomrule
  \end{tabular}%
  }
\end{table}

\clearpage
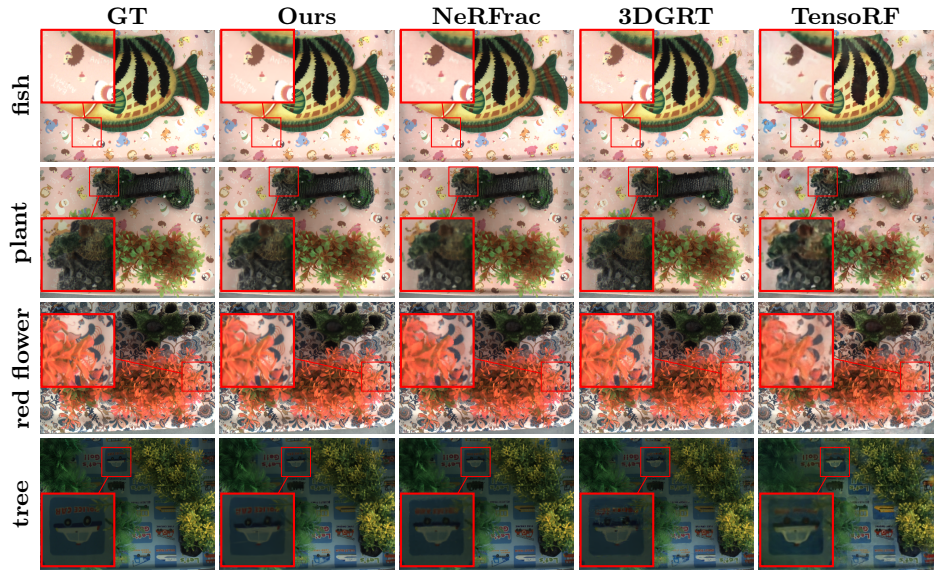
\begin{figure*}[htbp]
    \centering
    \setlength{\tabcolsep}{1pt}
    \setlength{\imagewidth}{0.19\textwidth}
    \renewcommand{\arraystretch}{0.6}
    \newcommand{\formattedgraphicsfirst}[1]{%
      \begin{tikzpicture}[spy using outlines={rectangle, magnification=2.5, connect spies}]
        \node[anchor=south west, inner sep=0] at (0,0){\includegraphics[width=\imagewidth]{#1}};
        \spy [red,size=27.5pt] on (.27\imagewidth,.17\imagewidth) in node at (.22\imagewidth,.55\imagewidth);
        \end{tikzpicture}%
    }
    \newcommand{\formattedgraphicssecond}[1]{%
      \begin{tikzpicture}[spy using outlines={rectangle, magnification=2.5, connect spies}]
        \node[anchor=south west, inner sep=0] at (0,0){\includegraphics[width=\imagewidth]{#1}};
        \spy [red,size=27.5pt] on (.37\imagewidth,.67\imagewidth) in node at (.22\imagewidth,.25\imagewidth);
        \end{tikzpicture}%s
    }
    \newcommand{\formattedgraphicsthird}[1]{%
      \begin{tikzpicture}[spy using outlines={rectangle, magnification=2.5, connect spies}]
        \node[anchor=south west, inner sep=0] at (0,0){\includegraphics[width=\imagewidth]{#1}};
        \spy [red,size=27.5pt] on (.9\imagewidth,.33\imagewidth) in node at (.22\imagewidth,.48\imagewidth);
        \end{tikzpicture}%
    }
    \newcommand{\formattedgraphicsfourth}[1]{%
      \begin{tikzpicture}[spy using outlines={rectangle, magnification=2.5, connect spies}]
        \node[anchor=south west, inner sep=0] at (0,0){\includegraphics[width=\imagewidth]{#1}};
        \spy [red,size=27.5pt] on (.44\imagewidth,.62\imagewidth) in node at (.22\imagewidth,.23\imagewidth);
        \end{tikzpicture}%
    }
    \begin{tabular}{m{0.38cm}<{\centering}m{\imagewidth}<{\centering}m{\imagewidth}<{\centering}m{\imagewidth}<{\centering}m{\imagewidth}<{\centering}m{\imagewidth}<{\centering}}
        & \textbf{GT} & \textbf{Ours} & \textbf{NeRFrac} & \textbf{3DGRT} & \textbf{TensoRF} \\
    \rotatebox{90}{\textbf{fish}} & \formattedgraphicsfirst{figures/supp/results/GT/1_real_fish.png} & \formattedgraphicsfirst{figures/supp/results/Ours/1_real_fish.png} & \formattedgraphicsfirst{figures/supp/results/NeRFrac/1_real_fish.png} & \formattedgraphicsfirst{figures/supp/results/3DGRT/1_real_fish.png} & \formattedgraphicsfirst{figures/supp/results/TensoRF/1_real_fish.jpg} \\
    \rotatebox{90}{\textbf{plant}} & \formattedgraphicssecond{figures/supp/results/GT/1_real_plant.png} & \formattedgraphicssecond{figures/supp/results/Ours/1_real_plant.png} & \formattedgraphicssecond{figures/supp/results/NeRFrac/1_real_plant.png} & \formattedgraphicssecond{figures/supp/results/3DGRT/1_real_plant.png} & \formattedgraphicssecond{figures/supp/results/TensoRF/1_real_plant.jpg} \\
    \rotatebox{90}{\textbf{red flower}} & \formattedgraphicsthird{figures/supp/results/GT/1_real_redflower.png} & \formattedgraphicsthird{figures/supp/results/Ours/1_real_redflower.png} & \formattedgraphicsthird{figures/supp/results/NeRFrac/1_real_redflower.png} & \formattedgraphicsthird{figures/supp/results/3DGRT/1_real_redflower.png} & \formattedgraphicsthird{figures/supp/results/TensoRF/1_real_redflower.jpg} \\
    \rotatebox{90}{\textbf{tree}} & \formattedgraphicsfourth{figures/supp/results/GT/1_real_tree.png} & \formattedgraphicsfourth{figures/supp/results/Ours/1_real_tree.png} & \formattedgraphicsfourth{figures/supp/results/NeRFrac/1_real_tree.png} & \formattedgraphicsfourth{figures/supp/results/3DGRT/1_real_tree.png} & \formattedgraphicsfourth{figures/supp/results/TensoRF/1_real_tree.jpg} \\
    \end{tabular}
    \vspace*{-4mm}
    \caption{\label{fig:supp:nerfracreal} 
    Qualitative results on the Nerfrac real dataset.}
    \vspace*{-6mm}
\end{figure*}

\begin{figure*}[htbp]
    \centering
    \setlength{\tabcolsep}{1pt}
    \setlength{\imagewidth}{0.19\textwidth}
    \renewcommand{\arraystretch}{0.6}
    \newcommand{\formattedgraphicsfirst}[1]{%
      \begin{tikzpicture}[spy using outlines={rectangle, magnification=2.5, connect spies}]
        \node[anchor=south west, inner sep=0] at (0,0){\includegraphics[width=\imagewidth]{#1}};
        \spy [red,size=27.5pt] on (.9\imagewidth,.2\imagewidth) in node at (.22\imagewidth,.48\imagewidth);
        \end{tikzpicture}%
    }
    \newcommand{\formattedgraphicssecond}[1]{%
      \begin{tikzpicture}[spy using outlines={rectangle, magnification=2.5, connect spies}]
        \node[anchor=south west, inner sep=0] at (0,0){\includegraphics[width=\imagewidth]{#1}};
        \spy [red,size=27.5pt] on (.7\imagewidth,.3\imagewidth) in node at (.22\imagewidth,.48\imagewidth);
        \end{tikzpicture}%
    }
    \newcommand{\formattedgraphicsthird}[1]{%
      \begin{tikzpicture}[spy using outlines={rectangle, magnification=2.5, connect spies}]
        \node[anchor=south west, inner sep=0] at (0,0){\includegraphics[width=\imagewidth]{#1}};
        \spy [red,size=27.5pt] on (.83\imagewidth,.37\imagewidth) in node at (.22\imagewidth,.48\imagewidth);
        \end{tikzpicture}%
    }
    \begin{tabular}{m{0.38cm}<{\centering}m{\imagewidth}<{\centering}m{\imagewidth}<{\centering}m{\imagewidth}<{\centering}m{\imagewidth}<{\centering}m{\imagewidth}<{\centering}}
    & \textbf{GT} & \textbf{Ours} & \textbf{NeRFrac} & \textbf{3DGRT} & \textbf{TensoRF} \\
    \rotatebox{90}{\textbf{hybrid}} & \formattedgraphicsfirst{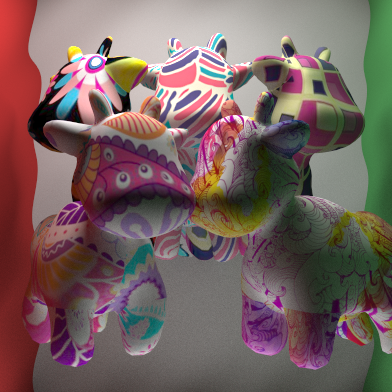} & \formattedgraphicsfirst{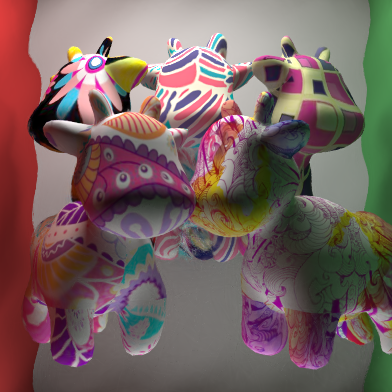} & \formattedgraphicsfirst{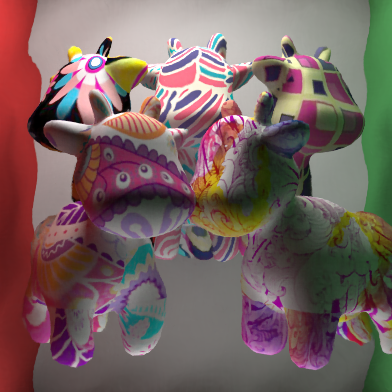} & \formattedgraphicsfirst{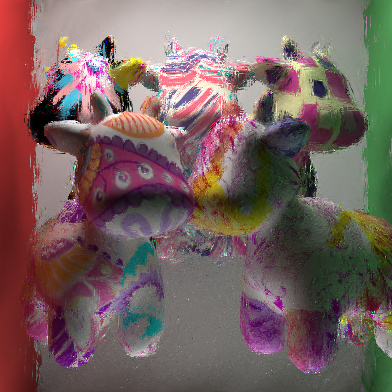} & \formattedgraphicsfirst{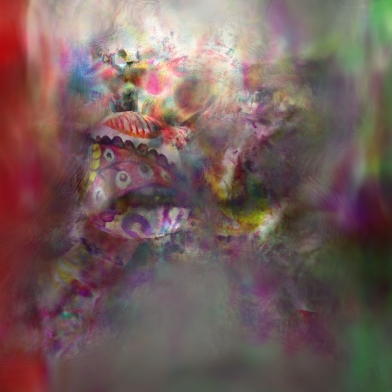} \\
    \rotatebox{90}{\textbf{primary sine}} & \formattedgraphicssecond{figures/supp/results/GT/2_primary.png} & \formattedgraphicssecond{figures/supp/results/Ours/2_primary.png} & \formattedgraphicssecond{figures/supp/results/NeRFrac/2_primary.png} & \formattedgraphicssecond{figures/supp/results/3DGRT/2_primary.png} & \formattedgraphicssecond{figures/supp/results/TensoRF/2_primary.jpg} \\
    \rotatebox{90}{\textbf{second sine}} & \formattedgraphicsthird{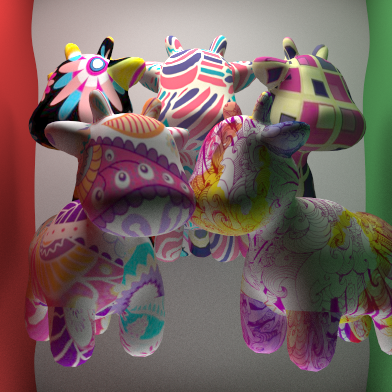} & \formattedgraphicsthird{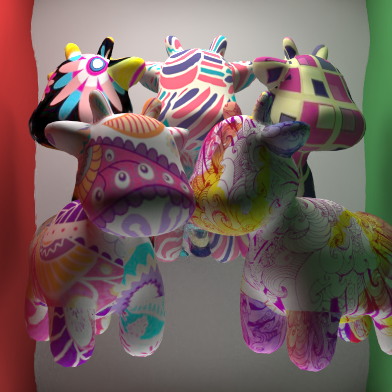} & \formattedgraphicsthird{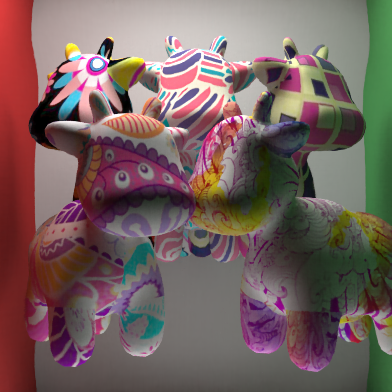} & \formattedgraphicsthird{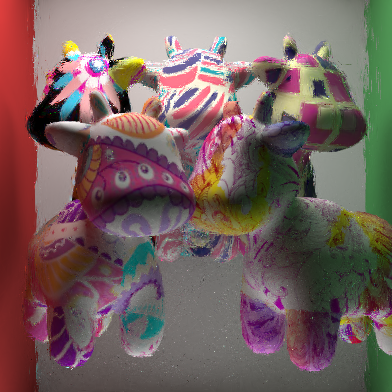} & \formattedgraphicsthird{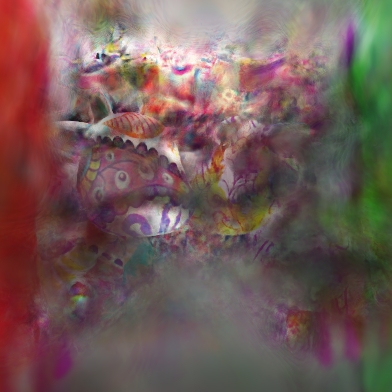} \\
    
    \end{tabular}
    \vspace*{-4mm}
    \caption{\label{fig:supp:nerfracsyn} 
    Qualitative results on the Nerfrac synthetic dataset.}
    \vspace*{-25mm}
\end{figure*}

\begin{figure*}[htbp]
    \centering
    \setlength{\tabcolsep}{1pt}
    \setlength{\imagewidth}{0.19\textwidth}
    \renewcommand{\arraystretch}{0.6}
    \newcommand{\formattedgraphicsfirst}[1]{%
      \begin{tikzpicture}[spy using outlines={rectangle, magnification=2.5, connect spies}]
        \node[anchor=south west, inner sep=0] at (0,0){\includegraphics[width=\imagewidth]{#1}};
        \spy [red,size=27.5pt] on (.37\imagewidth,.67\imagewidth) in node at (.22\imagewidth,.25\imagewidth);
        \end{tikzpicture}%
    }
    \newcommand{\formattedgraphicssecond}[1]{%
      \begin{tikzpicture}[spy using outlines={rectangle, magnification=2.5, connect spies}]
        \node[anchor=south west, inner sep=0] at (0,0){\includegraphics[width=\imagewidth]{#1}};
        \spy [red,size=27.5pt] on (.68\imagewidth,.63\imagewidth) in node at (.22\imagewidth,.48\imagewidth);
        \end{tikzpicture}%
    }
    \newcommand{\formattedgraphicsthird}[1]{%
      \begin{tikzpicture}[spy using outlines={rectangle, magnification=2.5, connect spies}]
        \node[anchor=south west, inner sep=0] at (0,0){\includegraphics[width=\imagewidth]{#1}};
        \spy [red,size=27.5pt] on (.8\imagewidth,.7\imagewidth) in node at (.22\imagewidth,.48\imagewidth);
        \end{tikzpicture}%
    }
    \begin{tabular}{m{0.38cm}<{\centering}m{\imagewidth}<{\centering}m{\imagewidth}<{\centering}m{\imagewidth}<{\centering}m{\imagewidth}<{\centering}m{\imagewidth}<{\centering}}
    & \textbf{GT} & \textbf{Ours} & \textbf{NeRFrac} & \textbf{3DGRT} & \textbf{TensoRF} \\
    \rotatebox{90}{\textbf{desktop}} & \formattedgraphicsfirst{figures/supp/results/GT/4_desktop.png} & \formattedgraphicsfirst{figures/supp/results/Ours/4_desktop.png} & \formattedgraphicsfirst{figures/supp/results/NeRFrac/4_desktop.png} & \formattedgraphicsfirst{figures/supp/results/3DGRT/4_desktop.png} & \formattedgraphicsfirst{figures/supp/results/TensoRF/4_desktop.jpg} \\
    \rotatebox{90}{\textbf{kitchen}} & \formattedgraphicssecond{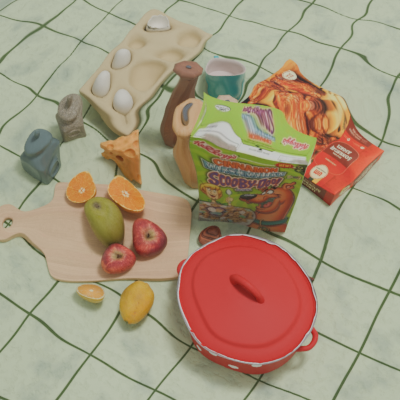} & \formattedgraphicssecond{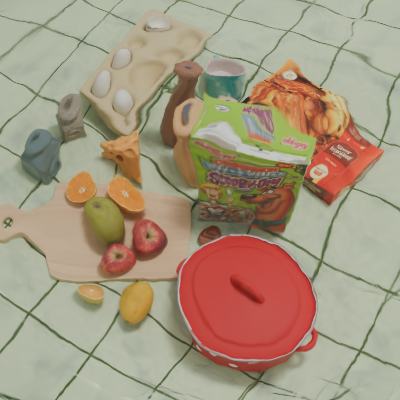} & \formattedgraphicssecond{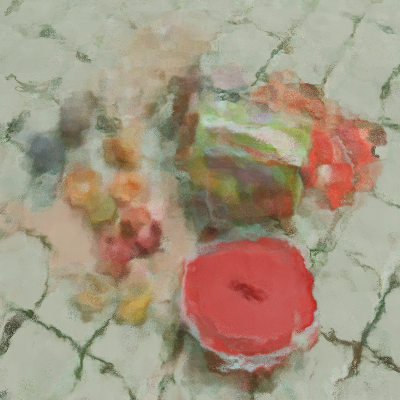} & \formattedgraphicssecond{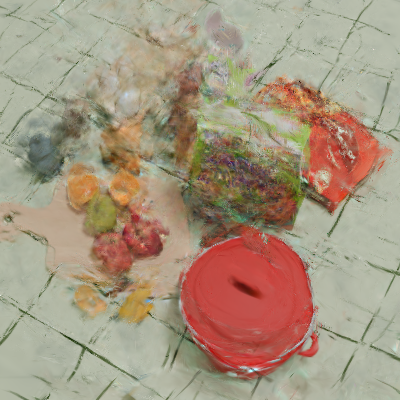} & \formattedgraphicssecond{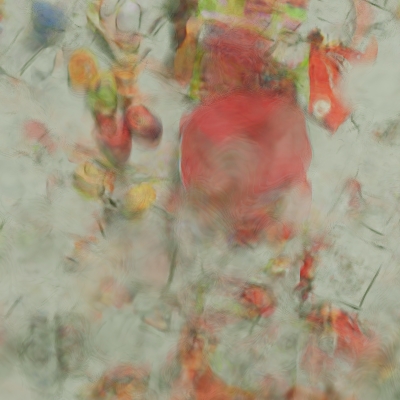} \\
    \rotatebox{90}{\textbf{toys}} & \formattedgraphicsthird{figures/supp/results/GT/4_toys.png} & \formattedgraphicsthird{figures/supp/results/Ours/4_toys.png} & \formattedgraphicsthird{figures/supp/results/NeRFrac/4_toys.png} & \formattedgraphicsthird{figures/supp/results/3DGRT/4_toys.png} & \formattedgraphicsthird{figures/supp/results/TensoRF/4_toys.jpg} \\
    
    \end{tabular}
    \vspace*{-3mm}
    \caption{\label{fig:supp:refracgs} 
    Qualitative results on the RefracGS dataset.}
    \vspace*{-3mm}
\end{figure*}

\begin{figure*}[htbp]
    \centering
    \setlength{\tabcolsep}{1pt}
    \setlength{\imagewidth}{0.19\textwidth}
    \renewcommand{\arraystretch}{0.6}
    \newcommand{\formattedgraphicsfirst}[1]{%
      \begin{tikzpicture}[spy using outlines={rectangle, magnification=2.5, connect spies}]
        \node[anchor=south west, inner sep=0] at (0,0){\includegraphics[width=\imagewidth]{#1}};
        \spy [red,size=27.5pt] on (.62\imagewidth,.85\imagewidth) in node at (.22\imagewidth,.48\imagewidth);
        \end{tikzpicture}%
    }
    \newcommand{\formattedgraphicssecond}[1]{%
      \begin{tikzpicture}[spy using outlines={rectangle, magnification=2.5, connect spies}]
        \node[anchor=south west, inner sep=0] at (0,0){\includegraphics[width=\imagewidth]{#1}};
        \spy [red,size=27.5pt] on (.51\imagewidth,.15\imagewidth) in node at (.22\imagewidth,.60\imagewidth);
        \end{tikzpicture}%
    }
    \newcommand{\formattedgraphicsthird}[1]{%
      \begin{tikzpicture}[spy using outlines={rectangle, magnification=2.5, connect spies}]
        \node[anchor=south west, inner sep=0] at (0,0){\includegraphics[width=\imagewidth]{#1}};
        \spy [red,size=27.5pt] on (.8\imagewidth,.4\imagewidth) in node at (.22\imagewidth,.48\imagewidth);
        \end{tikzpicture}%
    }
    \begin{tabular}{m{0.38cm}<{\centering}m{\imagewidth}<{\centering}m{\imagewidth}<{\centering}m{\imagewidth}<{\centering}m{\imagewidth}<{\centering}m{\imagewidth}<{\centering}}
    & \textbf{GT} & \textbf{Ours} & \textbf{NeRFrac} & \textbf{3DGRT} & \textbf{TensoRF} \\
    \rotatebox{90}{\textbf{desktop}} & \formattedgraphicsfirst{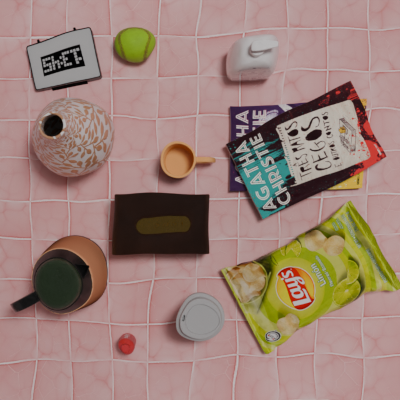} & \formattedgraphicsfirst{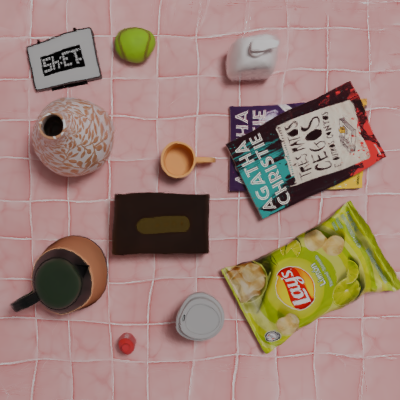} & \formattedgraphicsfirst{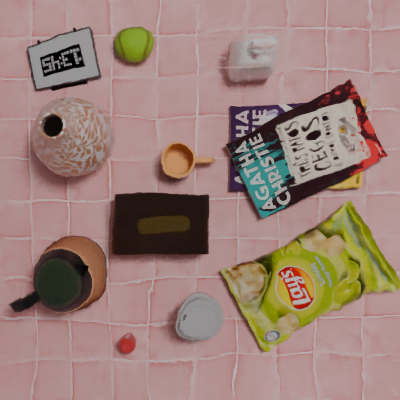} & \formattedgraphicsfirst{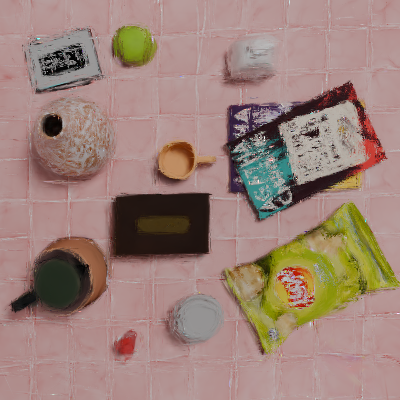} & \formattedgraphicsfirst{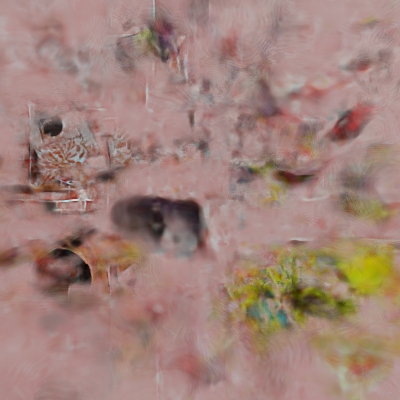} \\
    \rotatebox{90}{\textbf{kitchen}} & \formattedgraphicssecond{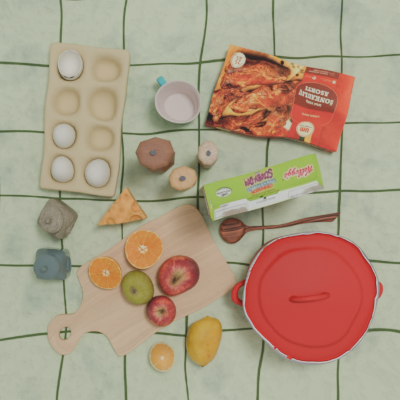} & \formattedgraphicssecond{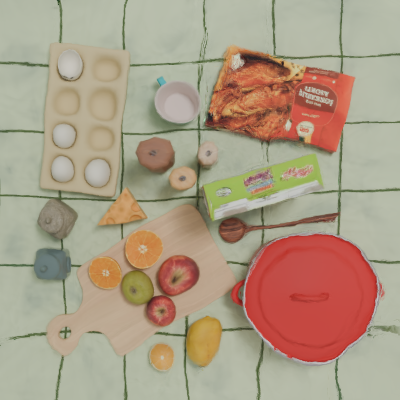} & \formattedgraphicssecond{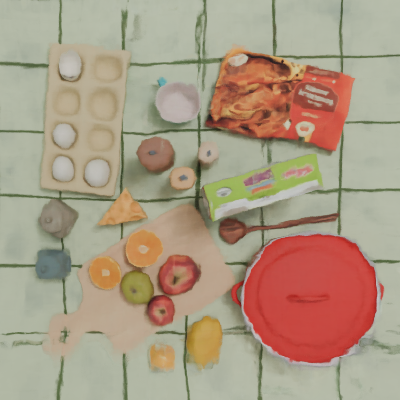} & \formattedgraphicssecond{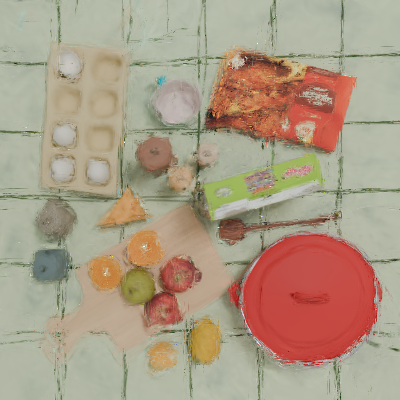} & \formattedgraphicssecond{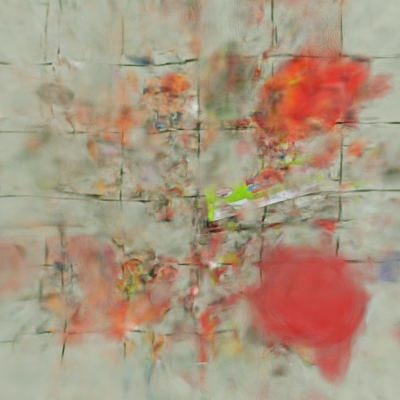} \\
    \rotatebox{90}{\textbf{toys}} & \formattedgraphicsthird{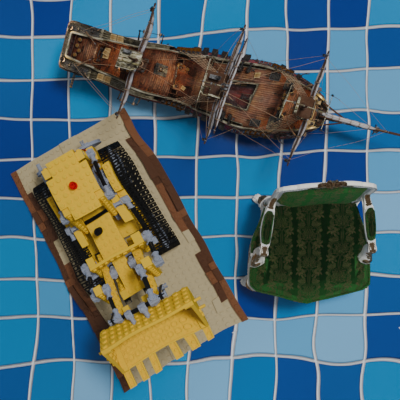} & \formattedgraphicsthird{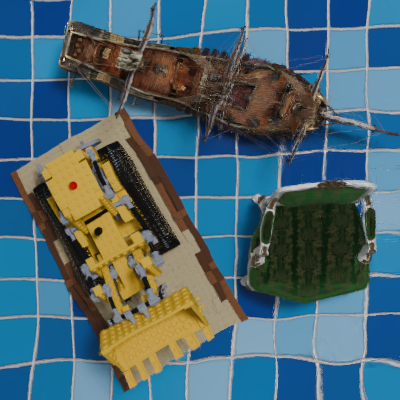} & \formattedgraphicsthird{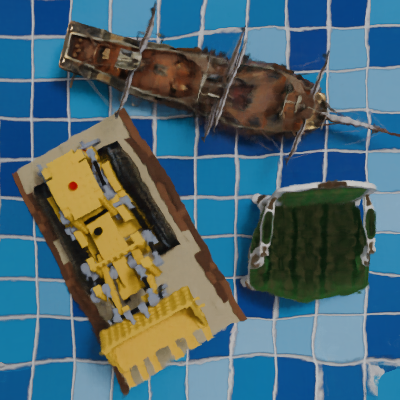} & \formattedgraphicsthird{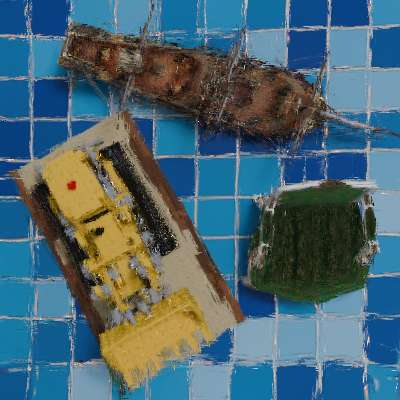} & \formattedgraphicsthird{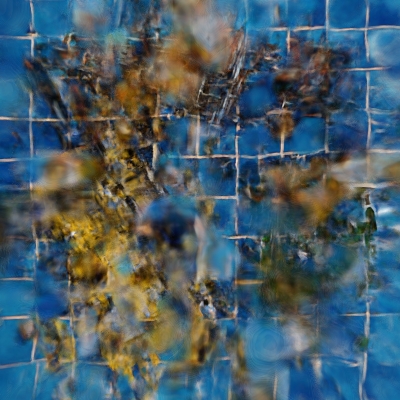} \\
    
    \end{tabular}
    \vspace*{-3mm}
    \caption{\label{fig:supp:refracgs9} 
    Qualitative results of different methods on RefracGS dataset scenes under the additional nine-view setting as discussed in \sec~\ref{sec:supp:camera9}.}
    \vspace*{-3mm}
\end{figure*}